%% file: main.tex
\documentclass[11pt]{article}

\usepackage[final]{acl}

\usepackage{algorithm}
\usepackage{algpseudocode}
\usepackage[most]{tcolorbox}
\usepackage{amssymb}
\usepackage{makecell}

\usepackage{xcolor}

 \usepackage{microtype}

\usepackage{hyperref}

\usepackage[english,bidi=default]{babel} 
\babelfont{rm}{TeXGyreTermesX} 
\babelprovide[import]{hindi}
\babelprovide[import]{japanese}
\babelfont[*japanese]{rm}{HaranoAjiMincho-Regular.otf}
\babelfont[*devanagari]{rm}{Lohit Devanagari}
\babelprovide[import]{arabic}
\babelfont[*arabic]{rm}{Noto Sans Arabic}

\usepackage{latexsym}

\usepackage{microtype}

\usepackage{tikz}
\usepackage{amsmath}
\usepackage{xcolor}
\usetikzlibrary{
  positioning,
  shapes.geometric,
  arrows.meta,
  fit,
  calc,
  decorations.pathreplacing,
  backgrounds
}
 
\definecolor{myblue}{RGB}{68,114,196}
\definecolor{mygreen}{RGB}{112,173,71}
\definecolor{myyellow}{RGB}{255,217,102}
\definecolor{mypurple}{RGB}{155,129,200}
\definecolor{myorange}{RGB}{237,125,49}
\definecolor{myred}{RGB}{192,0,0}
\definecolor{mypink}{RGB}{240,176,169}
\definecolor{mylightblue}{RGB}{189,215,238}
\definecolor{encodergreen}{RGB}{169,209,142}
\definecolor{decoderpink}{RGB}{240,176,169}

\newcommand{\flame}[3]{%
  \node at (#1,#2) {\includegraphics[scale=#3]{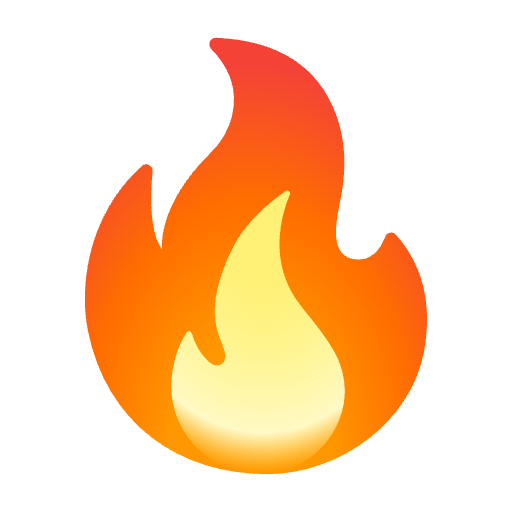}};
}

\newcommand{\snowflake}[3]{%
  \node at (#1,#2) {\includegraphics[scale=#3]{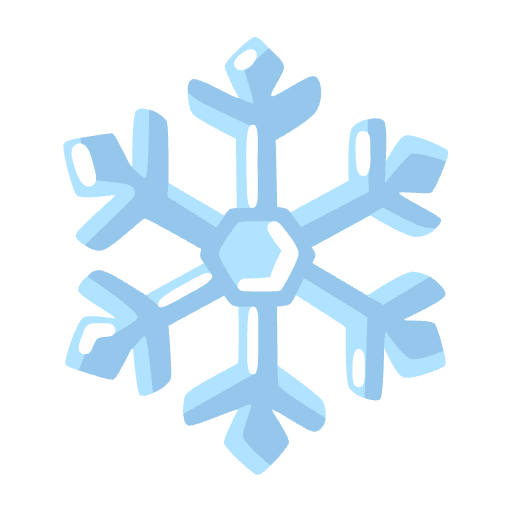}};
}

\newcommand{\salt}[1]{%
  \raisebox{-0.11em}{\includegraphics[scale=#1]{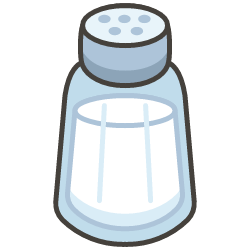}}%
}

\newcommand{\salttikz}[3]{ \node at (#1,#2) {\includegraphics[scale=#3]{latex/figs/salt.png}}; }

\usepackage{inconsolata}

\usepackage{graphicx}
\usepackage{todonotes}
\usepackage{booktabs}
\usepackage{multirow}

\usepackage{listings}
\title{Improving Cross-Lingual Token Representations by \\ Adding a Pinch of SALT \salt{0.05}}

\author{Guillem Ramírez Santos \\
  ILCC, University of Edinburgh \\
\texttt{	gramirez@ed.ac.uk} }

\begin{document}
\maketitle
\begin{abstract}
\looseness=-1

Cross-lingual sentence encoders enable scalable transfer across hundreds of languages, powering applications such as translation mining and zero-shot learning in low-resource settings. Although trained for sentence-level alignment, they are increasingly also applied to token-level tasks such as hallucination detection and sequence tagging, exposing a mismatch between training and usage. We propose SALT, a lightweight post-training method that improves token representations by injecting span-level supervision into existing sentence encoders. Across five multilingual token-level benchmarks, SALT achieves the best overall results on four of them, outperforming alternative fine-tuning strategies and competitive encoders. It also improves sentence-level performance on cross-lingual retrieval and classification tasks. These results demonstrate that span-level supervision is an effective signal for improving both token and sentence representations.
\end{abstract}

\input{latex/1_intro}
\input{latex/2_related_work}

\input{latex/3_method}

\input{latex/4_experiments}
\input{latex/5_results}
\input{latex/6_conclusions}
\newpage
\input{latex/99_ethics}

\bibliography{custom}

\appendix

\label{sec:appendix}
\input{latex/98_appendix}

\end{document}

%% file: latex/1_intro.tex
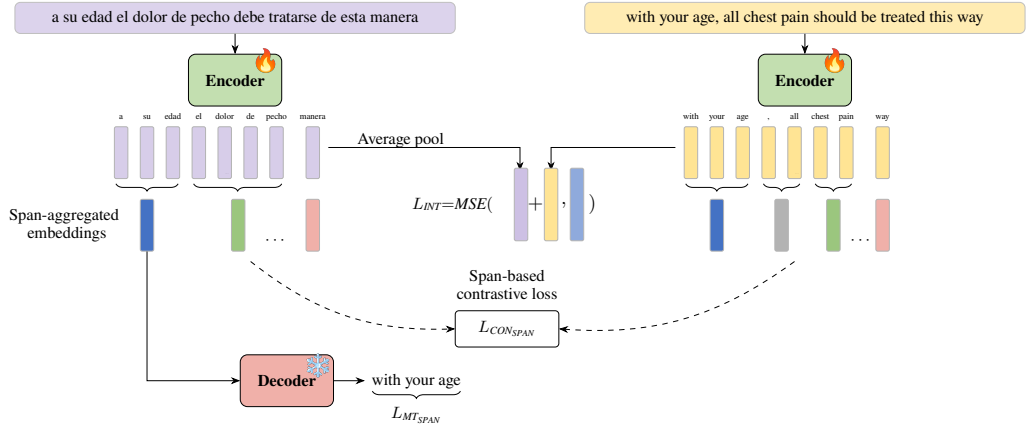
\begin{figure*}[t]
  \centering
  \resizebox{0.9\textwidth}{!}{%
    \begin{tikzpicture}[
      >=Stealth, font=\small,
      encoder/.style={draw,rounded corners=4pt,fill=encodergreen!70,minimum width=1.8cm,minimum height=1.0cm,align=center,font=\small\bfseries},
      decoder/.style={draw,rounded corners=4pt,fill=decoderpink,minimum width=1.8cm,minimum height=1.0cm,align=center,font=\small\bfseries},
      loss/.style={draw,rounded corners=2pt,fill=white,minimum width=2.0cm,minimum height=0.7cm,align=center,font=\small},
      emb/.style={draw=gray!60,rounded corners=1pt,fill=#1,minimum width=0.26cm,minimum height=1.4cm},
      embsm/.style={draw=gray!60,rounded corners=1pt,fill=#1,minimum width=0.26cm,minimum height=1.0cm},
      sent/.style={draw=gray!50,rounded corners=3pt,fill=#1,minimum height=0.6cm,inner xsep=5pt,font=\small,align=center}
    ]

    \node[font=\large\bfseries, anchor=west] at (-9.5, -0.3) {Pre-processing};
    \node[font=\small, anchor=center] at (0.5, -0.7) {Span alignment extraction};

    \node[sent=myyellow!50, minimum width=8.5cm] (en-full) at (-4.0, -1.4)
      {with your age, all chest pain should be treated this way};
    \node[sent=mypurple!35, minimum width=8.5cm] (es-full) at (-4.0, -2.3)
      {a su edad el dolor de pecho debe tratarse de esta manera};

    \draw[->, thick] (0.30, -1.85) -- (1.05, -1.85);

    \node[sent=myblue, text=white, anchor=west, minimum width=2.2cm] at (1.15,-1.4) {with your age};
    \node[sent=mylightblue, anchor=west, minimum width=0.9cm] at (3.53,-1.4) {, all};
    \node[sent=mygreen!60, anchor=west, minimum width=1.9cm] at (4.51,-1.4) {chest pain};
    \node[sent=myyellow!80, anchor=west, minimum width=2.9cm] at (6.49,-1.4) {should be treated};
    \node[sent=myorange!70, anchor=west, minimum width=1.4cm] at (9.47,-1.4) {this way};

    \node[sent=myblue, text=white, anchor=west, minimum width=1.8cm] at (1.15,-2.3) {a su edad};
    \node[sent=mygreen!60, anchor=west, minimum width=2.8cm] at (3.03,-2.3) {el dolor de pecho};
    \node[sent=myyellow!80, anchor=west, minimum width=2.3cm] at (5.91,-2.3) {debe tratarse};
    \node[sent=myorange!70, anchor=west, minimum width=2.3cm] at (8.29,-2.3) {de esta manera};

    \node[encoder] (enc-top) at (-2.5,-4.0) {Encoder};
    \snowflake{-2.5+0.6}{-3.57-0.15}{0.03}
\draw[->] (es-full.south) -- ++(0,-1.35) |- (enc-top.west);

    \draw[->] (enc-top.east) -- node[above,font=\small,align=center]{Sentence-level\\representation} (1.5,-4.0);

    \foreach \i in {1}{ \node[emb=myyellow!70] at (2.0+\i*0.30-0.30,-4.0) {}; }
    \node[font=\normalsize] at (2.32,-4.0) {$+$};
    \foreach \i in {1}{ \node[emb=mypurple!50] at (2.65+\i*0.30-0.30,-4.0) {}; }
    \node[font=\normalsize] at (2.97,-4.0) {$=$};
    \foreach \i in {1}{ \node[emb=myblue!60] at (3.3+\i*0.30-0.30,-4.0) {}; }
    \node[font=\small,align=left] at (4.4,-4.0) {Teacher\\embedding};

    \draw[gray,line width=0.6pt] (-9.5,-5.1) -- (11.5,-5.1);
    \node[font=\large\bfseries,anchor=west] at (-9.5,-5.6) {Training};
     \salttikz{-7.3}{-5.6}{0.05}


    \node[sent=mypurple!35,minimum width=8.5cm] (es-bot) at (-4.5,-6.5)
      {a su edad el dolor de pecho debe tratarse de esta manera};
    \node[sent=myyellow!50,minimum width=8.5cm] (en-bot) at (6.5,-6.5)
      {with your age, all chest pain should be treated this way};

    \node[encoder] (enc-bot-L) at (-4.5,-7.7) {Encoder};
    \flame{-4.82+0.9}{-7.33}{0.03}
    \node[encoder] (enc-bot-R) at (6.5,-7.7) {Encoder};
    \flame{6.18+0.9}{-7.33}{0.03}
    \draw[->] (es-bot.south) -- (enc-bot-L.north);
    \draw[->] (en-bot.south) -- (enc-bot-R.north);

    \foreach \i in {1,...,7}{ \node[embsm=mypurple!40] at (-7.2+\i*0.5,-9.1) {}; }

    \node[font=\tiny, scale=0.2] at ({-9.15+9*0.5},-9.5) {$\cdots$};
\foreach \word/\i in {a/1,su/2,edad/3,el/4,dolor/5,de/6,pecho/7}{
  \node[font=\tiny, scale=0.8, anchor=base] 
  at ({-7.2+\i*0.5},-8.45) {\word};}
\node[font=\tiny, scale=0.8, anchor=base] at (-3.0,-8.45) {manera};
    
    \draw[decorate,decoration={brace,amplitude=4pt,mirror}] (-6.80,-9.70) -- (-5.6,-9.70);

    \draw[decorate,decoration={brace,amplitude=4pt,mirror}] (-5.30,-9.70) -- (-3.6,-9.70);

    \foreach \i in {1}{\node[embsm=mypurple!40] at (-3.0,-9.1) {}; }
        

    \foreach \i in {1,...,7}{ \node[embsm=myyellow!70] at (3.8+\i*0.5,-9.1) {}; }
      \node[font=\tiny, scale=0.2] at ({-9.15+11+9*0.5},-9.5) {$\cdots$};
      
    \foreach \word/\i in {with/1,your/2,age/3,{,}/4,all/5,chest/6,pain/7}{
  \node[font=\tiny, scale=0.8, anchor=base] 
  at ({-7.2+11+\i*0.5},-8.45) {\word};}
\node[font=\tiny, scale=0.8, anchor=base] at (-3.0+11,-8.45) {way};

    \draw[decorate,decoration={brace,amplitude=4pt,mirror}] (-6.80+11,-9.70) -- (-5.6+11,-9.70);

    \draw[decorate,decoration={brace,amplitude=4pt,mirror}] (5.7,-9.70) -- (6.4,-9.70);

    \draw[decorate,decoration={brace,amplitude=4pt,mirror}] (6.7,-9.70) -- (7.4,-9.70);

    \foreach \i in {1}{\node[embsm=myyellow!70] at (-3.0+11,-9.1) {}; }

    \node[font=\small] (avgL) at (-1.3,-8.85) {Average pool};
    \draw[->] (-2.7, -9) -- (0+1-0.3, -9) -- (-0+1-0.3,-9.45);

    \node[font=\small] at (-1.30+1.0,-10.1) {$L_{\mathit{INT}}{=}\mathit{MSE}($};
    \foreach \i in {1}{ \node[emb=mypurple!50] at (1+0.02+\i*0.30-0.30-0.30,-10.1) {}; }
    \node[font=\normalsize] at (1-0.85+0.9,-10.1) {$,$};
    \foreach \i in {1}{ \node[emb=myblue!60] at (1+1.1+\i*0.30-0.30-0.7,-10.1) {}; }
    \node[font=\normalsize] at (1+1.4-0.7,-10.1) {$)$};

    \node[embsm=myblue]     (sp1) at (-6.2,-10.5) {};
    \node[embsm=mygreen!70] (sp2) at (-4.44,-10.5) {};
    \node[font=\small] at (-3.7,-10.8) {$\cdots$};
    \node[embsm=mypink]     (sp3) at (-3.0,-10.5) {};
    \node[align=left,font=\small] at (-7.8,-10.5) {Span-aggregated\\\hspace{10pt}embeddings};

    \node[embsm=myblue]     (ep1) at (-6.2+11,-10.5) {};
    \node[embsm=gray!60]    (ep2) at (6.05,-10.5) {};
    \node[font=\small] at (-3.7+11.3,-10.8) {$\cdots$};
    \node[embsm=mygreen!70] (ep3) at (7.05,-10.5) {};
        \node[embsm=mypink]     (sp3) at (-3.0+11,-10.5) {};

    \node[font=\small,align=center] at (0.75,-11.65) {Span-based\\contrastive loss};
    \node[loss] (lcspan) at (0.75,-12.5) {$L_{\mathit{CON}_{\mathit{SPAN}}}$};
    \draw[->, dashed, bend right=20] (-4.2, -11.2) to (lcspan.west);
    \draw[->, dashed, bend right=-20] (-4.2+10.5, -11.2) to (lcspan.east);

    \node[decoder] (decoder) at (-3.5,-13.5) {Decoder};
    \snowflake{-3.5+0.6}{-13.0-0.2}{0.03}
    \draw[->] (sp1.south) -- (-6.20,-13.5) -- (decoder.west);
    \draw[->] (decoder.east) -- (-2.0,-13.5);
    \node[font=\small] at (-1.0,-13.5) {with your age};
    \draw[decorate,decoration={brace,amplitude=4pt,mirror}]
      (-1.85,-13.7) -- (-0.15,-13.7)
      node[midway,below=5pt,font=\small] {$L_{\mathit{MT}_{\mathit{SPAN}}}$};

    \end{tikzpicture}%
  }
  \caption{\textbf{Pre-processing} (top):
extraction of aligned spans and teacher embeddings from a pre-trained encoder.
           \textbf{Training SALT} (bottom): we train the encoder with span-level contrastive and translation losses to improve token-level representations, besides an interpolation loss that preserves sentence embeddings.}

           \vspace{-5pt}
  \label{diagram}
\end{figure*}

\section{Introduction}

Cross-lingual sentence encoders have become a core building block of multilingual NLP. Models such as SONAR~\citep{sonar} or MEXMA~\citep{mexma} learn a shared embedding space in which semantically equivalent sentences across languages are mapped to nearby representations, leading to strong performance on sentence-level tasks such as cross-lingual retrieval, sentence mining, and zero-shot classification.

Despite being trained for cross-lingual sentence alignment, these models are increasingly used as \emph{token-level representations} in downstream applications, including word alignment~\citep{miao-etal-2024-enhancing}, sequence tagging~\citep{mehta-varma-2023-llm}, hallucination detection~\citep{ottawa}, sentence segmentation~\citep{omnilingualmtteam2026omnilingualmtmachinetranslation} and label projection~\citep{labelproj}. This practice exposes a structural mismatch: sentence encoders are optimised to preserve global semantic similarity, yet are repurposed to support fine-grained token-level alignment. As a result, they may yield well-structured sentence embeddings but an inconsistent or noisy token representation.

A promising way to address this mismatch is to introduce training signals at a finer granularity by identifying aligned words across languages and incorporating token-level losses~\citep{alqahtani-etal-2021-using-optimal, miao-etal-2024-enhancing}. However, word-level alignment is often too restrictive, as meaning is frequently distributed across multiple tokens.

To this end, we propose using \emph{spans}—contiguous sequences of tokens that express equivalent meaning across languages. While this alignment signal is implicitly present in parallel data, previous sentence- or token-level objectives do not exploit it. We introduce \salt{0.04} \textbf{SALT} (\textbf{S}pan-\textbf{A}ligned \textbf{L}earning for cross-lingual \textbf{T}okens), a novel post-training method for multilingual sentence encoders (see Figure~\ref{diagram}). SALT augments standard sentence-level training with span-aligned contrastive and translation objectives that explicitly shape token- and span-level representations, while a sentence-level interpolation objective preserves the original embedding space. Importantly, SALT requires no architectural modifications and can be applied directly to existing encoders, while requiring substantially less training than multilingual pre-training.

We evaluate SALT on a suite of multilingual token-level benchmarks, including word alignment, cross-lingual sequence tagging, and word sense disambiguation. Across tasks, SALT consistently improves token-level performance while maintaining or improving sentence-level quality. In particular, it achieves the best overall performance among several fine-tuning strategies on four out of five token-level benchmarks, demonstrating that span-level supervision provides a strong complementary signal rather than a competing objective. To better understand these improvements, we analyse the geometry of the resulting token embedding spaces, examining isotropy, hubness, and language-specific clustering. We find that SALT yields better-distributed cross-lingual token representations. Our contributions are as follows:

\begin{itemize}


\item We propose \textbf{SALT}, a post-training method that injects span-level alignment into sentence encoders, bridging sentence-level training and token-level applications.\footnote{Model available at [OMIT]. Code available at \href{https://anonymous.4open.science/r/token_local-8E04}{https://anonymous.4open.science/r/token\_local-8E04}.}  

\item We demonstrate that span-aligned supervision consistently improves performance across multiple token-level multilingual tasks, while preserving and sometimes improving sentence-level embedding quality.

\item We analyse the geometry of the word embedding space and find that SALT yields a better-structured cross-lingual representation, improving hubness while preserving isotropy and reducing language-specific clustering.

\end{itemize}

%% file: latex/2_related_work.tex
\section{Related work}

\paragraph{Sentence encoders} Cross-lingual sentence encoders aim to learn a shared semantic space in which sentences with equivalent meaning across languages are mapped to nearby embeddings. Early approaches based on multilingual word embeddings have evolved into large-scale sentence encoders trained on parallel corpora using contrastive, regression, or decoding objectives. Models such as SONAR~\citep{sonar}, MEXMA~\citep{mexma}, OmniSONAR~\citep{sonar2} or LaBSE~\citep{labse} exemplify this paradigm. 


\paragraph{Word and phrase alignment}
Early statistical machine translation relied on word alignment models and derived phrase pairs as discrete translation units for decoding~\citep{DBLP:journals/coling/BrownPPM94, DBLP:conf/naacl/KoehnOM03}, establishing a notion of cross-lingual span correspondence. While this is related in spirit to SALT, we use such span correspondences only as a training signal. More recently, word alignment has been revisited in neural encoders, with methods extracting alignments from contextual representations~\citep{simalign}. Related work improves cross-lingual alignment through objectives that encourage finer-grained token-level correspondence~\citep{chi-etal-2021-improving, DBLP:conf/eacl/DouN21}.

\paragraph{Bridging sentence- and token-level representations}
Prior work improves cross-lingual sentence encoders by incorporating auxiliary word- or token-level objectives during pre-training~\citep{DBLP:conf/iclr/WeiW0XYL21, alqahtani-etal-2021-using-optimal, li-etal-2021-multi, dap, miao-etal-2024-enhancing}, typically relying on explicit alignment signals to enhance sentence-level embeddings. In contrast, we directly target token- and span-level representations, both in training and evaluation, explicitly focusing on improving their quality rather than treating them as a by-product of sentence-level learning, focusing on adapting existing sentence encoders for token-centric tasks. \citet{sonar2} propose OmniSONAR-Token, which extends SONAR with a word-alignment-based objective. SALT takes a different approach, combining contrastive and translation objectives to learn representations of contiguous multi-token spans. We show that SALT improves performance on token-level tasks and additionally evaluate its representations on span-level translation and retrieval.


%% file: latex/3_method.tex
\section{\salt{0.05} SALT: Span-Aligned Learning for Cross-Lingual Tokens}
To improve cross-lingual token representations produced by sentence encoders, we propose SALT, a training method that aligns semantically equivalent spans across languages. SALT can be applied post-training to any pre-trained encoder without modifying its architecture.
As illustrated in Figure~\ref{diagram}, we first extract semantically equivalent multi-word spans from parallel sentences and precompute their corresponding sentence representations. During training, span-level objectives refine token and span embeddings while preserving the structure of the original sentence-level representation space.

\subsection{Method} Let $x$ and $y$ be a translation pair. A tokeniser $\mathcal{T}$ with vocabulary $V$ maps them to token sequences $t_x=\mathcal{T} (x) \in V^n$ and $t_y=\mathcal{T} (y) \in V^m$. A pretrained encoder $\mathcal{E}$ then produces contextual embeddings $H_x = \mathcal{E}(t_x) \in \mathbb{R}^{n \times d}$ and $H_y = \mathcal{E}(t_y) \in \mathbb{R}^{m \times d}$. These representations are typically trained such that their pooled sentence embeddings are aligned, i.e., $\text{pool}(H_x) \approx \text{pool}(H_y)$. Following \citet{sonar}, we adopt average pooling, although our approach is compatible with alternative strategies, such as max pooling or the use of a designated [CLS] token (as in \citet{mexma}).

We define a \textit{span pair} as a pair of contiguous token subsequences from a translation pair with similar contextual meaning. Formally, it is given by $(t_x[i:j], t_y[i':j'])$. $t_x[i:j]$ denotes tokens $i,\dots,j−1$ with \(0\le i<j\le n;\;0\le i'<j'<m\). The span representations are obtained by pooling the token embeddings within each subsequence, i.e., $\text{pool}(H_x[i:j])$ and $\text{pool}(H_y[i':j'])$. 

\subsubsection{Span extraction}
We require aligned spans across parallel sentences to construct training supervision. We consider two approaches for span extraction: (i) an LLM-based method, and (ii) a lightweight heuristic, CASE.

In most of our experiments, we use LLaMA-70B~\citep{DBLP:journals/corr/abs-2407-21783} to extract aligned spans (see Appendix~\ref{span:extraction}). However, LLM-based extraction can be computationally expensive and not always available. To address this, we introduce CASE (Constituent Alignment Span Extraction), a heuristic that extracts aligned spans using token-level alignments derived from encoder $\mathcal{E}$ representations. 

\paragraph{CASE}
Given a source sentence $x$, we compute token-level alignments $M: \{0, \dots, n-1\} \rightarrow \{0, \dots, m-1\} \cup \{\varnothing\}$ using Argmax~\citep{simalign}, extended to multi-token words. For a target token $t$, we define the inverse mapping $M^{-1}(t) = \{ s \mid M(s) = t\}$.

We consider candidate source spans $t_x[i:j]$. Let $A_{i:j} = \{ M(s) \mid s \in \{i, \dots, j-1\},\; M(s) \neq \varnothing \}$. Then we define the aligned target span as the tightest span covering all aligned tokens:
$i' = \min A_{i:j}$, \;
$j' = \max A_{i:j} + 1$. 

We restrict source spans to constituents identified by the Benepar parser~\citep{DBLP:conf/acl/KitaevCK19}, as simpler heuristics often yielded less meaningful translation units during CASE development. We discard span pairs for which alignment links are not fully contained within the span pair, i.e., no token in the target span is aligned to a source token outside the source span, and vice versa.

Finally, we define the coverage of a span pair as the proportion of aligned tokens in both directions:
\begin{equation}
\begin{small}
\begin{aligned}
\mathrm{cover} &= \frac{\mid\{s \mid s \in \{i, \dots, j-1\}, M(s) \neq \varnothing \}\mid}{2((j - i)+(j' - i'))}  \\
    &+ \frac{\mid\{t  \mid  t \in \{i', \dots, j'-1\}, M^{-1}(t) \neq \varnothing \}\mid}{2((j - i)+(j' - i'))}
\end{aligned}
\end{small}
\end{equation}

We discard any pair of source-target spans whose coverage falls below a threshold $c$ = 0.7, ensuring that only well-aligned spans are used. 

\begin{figure}
  \centering
\includegraphics[trim={0cm 0cm 0cm 0cm},clip, width=\columnwidth]{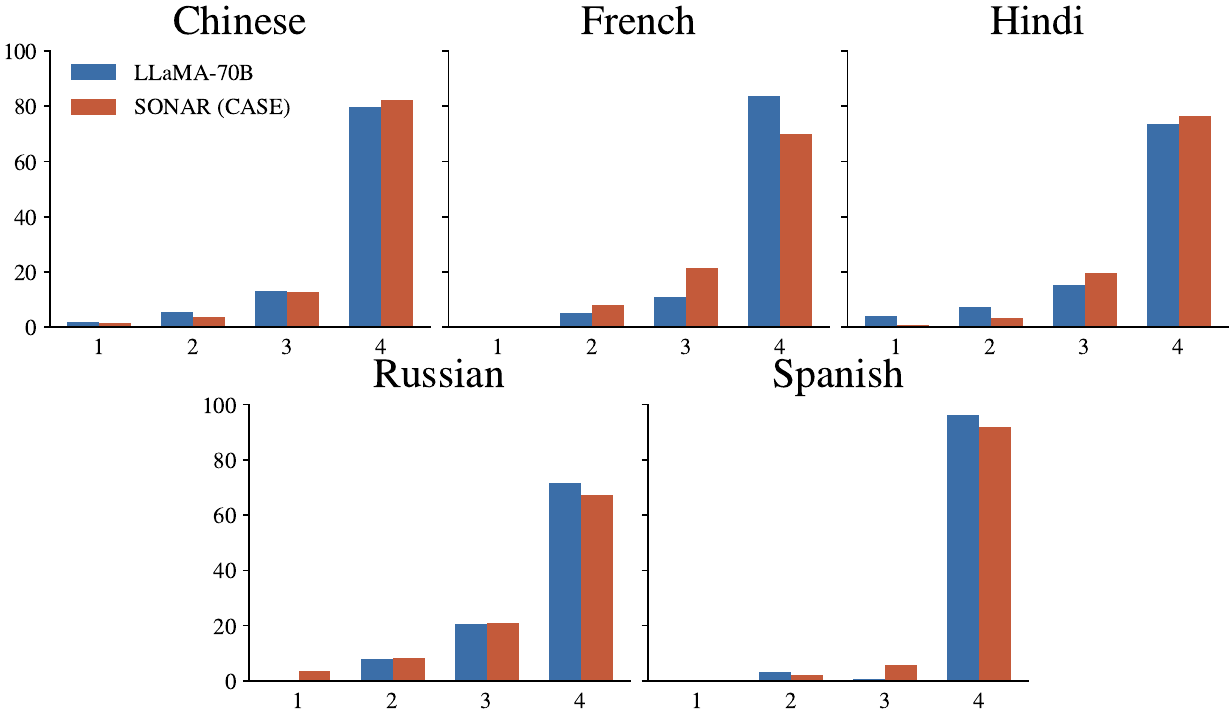}
    \caption{Human evaluation of extracted spans. Score distributions for LLaMA-70B and SONAR (CASE). }
    \label{human_eval}
    \vspace{-5pt}
\end{figure}

\paragraph{Evaluation of span extraction}
We evaluate span quality via human evaluation on Chinese, French, Hindi, Russian, and Spanish. Spans are extracted from English–target sentence pairs from our word alignment dataset (Section~\ref{token_eval}). We recruit 7 native speakers to annotate a total of 2,361 span pairs for both LLaMA-70B and CASE (with SONAR), rating span semantic equivalence on a 4-point scale: (1) unrelated, (2) somewhat related, (3) nearly equivalent, (4) equivalent. Figure~\ref{human_eval} shows the score distribution. Both LLaMA-based extraction and CASE yield high-quality span pairs, with over 90\% of annotations in the top two categories. LLaMA obtains more top scores (4) for Spanish, Russian, and French, while CASE matches or exceeds it in other languages. We retain score-4 spans to construct CrossSpan for the evaluation in Section~\ref{analysis_sec}. Overall, both methods effectively identify semantically equivalent spans, and CASE offers an efficient alternative to LLM-based extraction. See Appendix~\ref{annotation} for annotation guidelines and full results.

\subsubsection{SALT training objective} SALT uses a translation (MT) and a contrastive (CON) objective at the span level, as well as an interpolation objective (INT) at the sentence level.
\vspace{-2pt}
\begin{equation}
    \mathcal{L}_{SALT} =  \mathcal{L}_{CON_{Span}} + \alpha \mathcal{L}_{MT_{Span}} + \beta \mathcal{L}_{INT} 
\end{equation}
where $\alpha, \beta \in \mathbb{R}$ are hyperparameters set to one unless otherwise specified. For simplicity, we present the losses for a single translation pair $(x, y)$, though they are computed in batches during training.
\paragraph{Contrastive objective} 
We apply the span extraction method (LLM or CASE) to obtain a collection of $p$ aligned spans 
$t_x[i^1: j^1], \dots, t_x[i^p: j^p]$ and $t_y[i'^1: j'^1],\dots, t_y[i'^p: j'^p]$.

To increase coverage of both sentences, we augment this set with unaligned spans. Specifically, for every maximal contiguous subsequence of tokens that does not belong to any aligned span, we form an additional span. This yields additional spans $(t_x[i^{p+1}: j^{p+1}], \dots, t_x[i^{q}: j^{q}])$ and $(t_y[i'^{p+1}: j'^{p+1}], \dots, t_y[i'^{v}: j'^{v}])$.

We compute the corresponding span embeddings $H_x^1,\dots,H_x^q$ and $H_y^1,\dots,H_y^v$ by averaging the token embeddings within each span. We then define the contrastive loss as

\begin{equation}
\begin{small}
\begin{aligned}
 \mathcal{L}_{\text{CON}_{Span}} 
 &= -\frac{1}{p} \sum_{i=1}^{p} \Bigg[
\log
\frac{\exp(\langle H_x^i, H_y^i \rangle / \tau)}
{\sum_{j=1}^{v} \exp(\langle H_x^i, H_y^j \rangle / \tau)}
\\
&\qquad +
\log
\frac{\exp(\langle H_x^i, H_y^i \rangle / \tau)}
{\sum_{j=1}^{q} \exp(\langle H_x^j, H_y^i \rangle / \tau)}
\Bigg]
\end{aligned}
\end{small}
\end{equation}

where $\langle \cdot, \cdot \rangle$ denotes the dot product, $(H_x^i, H_y^i)$ corresponds to an aligned span pair, and $\tau$ is a temperature hyperparameter. The first term applies a row-wise softmax over target spans for each source span, while the second term applies a column-wise softmax over source spans for each target span. This loss encourages semantically equivalent spans to have similar embeddings while pushing apart spans with different meanings.

\paragraph{Translation objective} 
Given a pre-trained decoder $\mathcal{D}$ that we keep frozen when training SONAR (see Appendix~\ref{extra_encoders} for other encoders), we define the translation loss as the cross-entropy over the target span:
\begin{equation}
\begin{small}
\mathcal{L}_{\text{MT}_{\text{Span}}} \! = \! -\tfrac{1}{p}\sum_{l=1}^{p} \sum_{k=i'^l}^{j'^l\!-\!1}\! \log P_\mathcal{D}\big(t_y[k] ; t_y[i'^l:k] , H_x^l\big)
\end{small}
\end{equation}
where ($t_x[i^l:j^l]$ and $t_y[i'^l:j'^l]$) is an extracted span pair, $H_x^l$ is the pooled source span representation and $P_\mathcal{D}$ denotes the probability assigned by the decoder to the next target token, conditioned on the previous target tokens and the pooled representation of the source span. 

\paragraph{Interpolation objective} 
Following~\citet{DBLP:conf/acl/TsiamasDC25}, we introduce a sentence-level interpolation objective that encourages the encoder to preserve the original sentence representation space. We obtain a teacher representation $z_{xy}$ by averaging the source and target sentence embeddings produced by the frozen encoder $\mathcal{E}^{frozen}$. The interpolation loss is then defined as

\begin{equation}
\small 
   \mathcal{L}_{INT} = \text{MSE}\Big(z_{xy}, \text{pool}(H_x)\Big)+\text{MSE}\Big(z_{xy}, \text{pool}(H_y)\Big)
\label{interpolation_loss}
\end{equation}

where $\text{MSE}$ denotes the mean squared error. Intuitively, this loss constrains the encoder so that token-level training objectives do not distort the sentence-level embedding space, maintaining consistency with the initial frozen encoder.

%% file: latex/4_experiments.tex
\begin{table}[t]
\centering
\resizebox{\columnwidth}{!}{%
\begin{tabular}{cc}
\toprule

\bf{\makecell{Task \\ (\#langs)}}& \textbf{Example} \\
\midrule
\makecell{AER\\(18)} &

\raisebox{-0.5\height}{%
\begin{tikzpicture}[
  every node/.style={inner sep=1pt, outer sep=0pt}
]

\node[anchor=base] (en1) at (0,0) {The};
\node[anchor=base] (en2) at ([xshift=8.1mm] en1.base east) {necessary};
\node[anchor=base] (en3) at ([xshift=9.1mm] en2.base east) {correction};
\node[anchor=base] (en4) at ([xshift=4.1mm] en3.base east) {will};
\node[anchor=base] (en5) at ([xshift=3.1mm] en4.base east) {be};
\node[anchor=base] (en6) at ([xshift=6.1mm] en5.base east) {made};
\node[anchor=base] (en7) at ([xshift=2.1mm] en6.base east) {.};

\node[anchor=base west] (de1) at ([yshift=-10mm] en1.base west) {Eine};
\node[anchor=base] (de2) at ([xshift=12mm] de1.base east) {entsprechende};
\node[anchor=base] (de3) at ([xshift=8.1mm] de2.base east) {Anderung};
\node[anchor=base] (de4) at ([xshift=4.1mm] de3.base east) {wird};
\node[anchor=base] (de5) at ([xshift=11mm] de4.base east) {vorgenommen};
\node[anchor=base] (de6) at ([xshift=6.1mm] de5.base east) {werden};
\node[anchor=base] (de7) at ([xshift=4.1mm] de6.base east) {.};

\draw[<->,thick] (de1.north) -- (en1.south);
\draw[<->,thick] (de2.north) -- (en2.south);
\draw[<->,thick] (de3.north) -- (en3.south);
\draw[<->,thick] (de4.north) -- (en4.south);
\draw[<->,thick] (de5.north) -- (en6.south);
\draw[<->,thick] (de6.north) -- (en5.south);
\draw[<->,thick] (de7.north) -- (en7.south);

\end{tikzpicture}%
}
\\
\addlinespace[8pt]

\makecell{Massive\\(52)} &
\begin{tabular}[c]{@{}l@{}}
What is today 's forecast for Berlin\\
        \makebox[0pt][l]{\hspace{3.7em}\footnotesize DATE}%
        \makebox[0pt][l]{\hspace{11.7em}\footnotesize	 PLACE}%
\end{tabular}
\\
\addlinespace[8pt]

\makecell{PAN-X\\(40)} &
\begin{tabular}[c]{@{}l@{}}
REDIRECCIÓN Algarrobo ( Chile ) \\

        \makebox[0pt][l]{\hspace{8.3em}\footnotesize B-LOC}%
        \makebox[0pt][l]{\hspace{12em}\footnotesize	 I-LOC}%

\end{tabular}
\\
\addlinespace[8pt]

\makecell{UPOS\\(51)} &
\begin{tabular}[c]{@{}l@{}}
\foreignlanguage{japanese}{葬儀\hspace{1.3em}の\hspace{1.3em}最中\hspace{1.3em}です\hspace{1.2em}よ\hspace{1.8em}!}  \\

        \makebox[0pt][l]{\hspace{0em}\footnotesize NOUN}%
        \makebox[0pt][l]{\hspace{3.1em}\footnotesize	 ADP}%
        \makebox[0pt][l]{\hspace{5.5em}\footnotesize	 NOUN}%
        \makebox[0pt][l]{\hspace{9.0em}\footnotesize	 AUX}%
        \makebox[0pt][l]{\hspace{11.6em}\footnotesize	 PART}%
        \makebox[0pt][l]{\hspace{14em}\footnotesize	 PUNCT}%

\end{tabular}
\\
\addlinespace[8pt]

\makecell{WiC \\(6)} &
\begin{tabular}[c]{@{}l@{}}
$\left.
\begin{array}{@{}l@{}}
\text{Bolivia holds a key \underline{play} in this process of peace.} \\
\text{A musical \underline{play} on the same subject was also staged.}
\end{array}
\right\}\ 
\begin{array}{@{}l@{}}
\text{Different}\\
\text{meaning}
\end{array}$
\end{tabular}
\\
\bottomrule
\end{tabular}}
\caption{Summary of the tasks and number of languages used for the evaluation of cross-lingual token representations. Source: \citet{sonar2}}
\label{example_datasets}
\vspace{-15pt}
\end{table}

\section{Experiment details}
We initialise $\mathcal{E}$,$\mathcal{D}$ with the respective pre-trained SONAR weights~\citep{sonar}.

\paragraph{Data} We fine-tune SONAR on a subset of the data originally used to train it, namely the publicly available NLLB Primary dataset~\citep{DBLP:journals/corr/abs-2207-04672}.~\footnote{Downloaded using the script provided at \href{https://github.com/gordicaleksa/Open-NLLB/blob/nllb_replication/examples/nllb/data/README.md}{https://github.com/gordicaleksa/Open-NLLB}.}
 This dataset comprises high-quality parallel sentences spanning a diverse set of languages. We begin by excluding all language pairs for which neither language belongs to the set of 57 designated test languages, yielding a total of 129 language pairs (see Appendix~\ref{langs_info}). The 29 test languages not observed during training can be used to evaluate the generalisation of our method (see Appendix~\ref{test_unseen_bruto}). 

For each retained language pair, we downsample the corpus to 40{,}000 sentence pairs, observing no degradation in performance as a result of this reduction. In line with the procedure of~\citet{sonar2}, we further refine the dataset by removing sentence pairs whose BLASER 2 score~\citep{DBLP:conf/emnlp/DaleC24} deviates by more than one standard deviation from the mean score computed for the given language pair.

\paragraph{Training hyperparameters} We train SONAR for 20,000 steps, with a maximum of 1400 tokens per batch (roughly corresponding to 50 pairs of sentences per batch) and a learning rate of $\mu=10e-5$. We use an AdamW optimiser~\citep{DBLP:conf/iclr/LoshchilovH19} and warm-up of 1000 steps. We do the training runs on three seeds and report average values in all experiments unless otherwise specified. Standard deviations are reported in Appendix~\ref{stdev}. Overall, this training is lightweight relative to pre-training, using over $3{,}500\times$ fewer tokens than SONAR
(Appendix~\ref{compute}).

\paragraph{Baselines} We report the performance of the encoders XLM-R~\citep{xlm}, XLM-Align~\citep{chi-etal-2021-improving}, LaBSE~\citep{labse}, and MEXMA~\citep{mexma}. In addition, we compare SALT against variants obtained by fine-tuning SONAR with alternative training objectives. Specifically, we train the encoder for longer using the original SONAR loss proposed by~\citet{sonar}. We also include token-level objectives: the Self-Objective (SO) loss~\citep{DBLP:conf/eacl/DouN21}, a contrastive word-alignment objective; the WordOT loss~\citep{alqahtani-etal-2021-using-optimal}, using Optimal Transport; the WACSE loss~\citep{miao-etal-2024-enhancing}, which adds a masked language modeling head; and the OmniSONAR-Token loss~\citep{sonar2}, which extends SO loss with interpolation. Unlike these word-level objectives, SALT additionally introduces span-level supervision through contrastive and translation objectives. See Appendix~\ref{losses:implement} for a more detailed description of the baselines as well as implementation details; we get the word alignments using Argmax for SO Loss, WACSE and OmniSONAR-Token. 

\paragraph{Sentence evaluation} We evaluate the task of sentence mining using the FLORES-200 devtest~\citep{nllb} (80 languages) and report xsim~\citep{labse} and xsim++~\citep{xsimplusplus}. For classification, we report on tasks from MTEB~\citep{muennighoff-etal-2023-mteb} (English only). 

\subsection{Cross-lingual token evaluation}
\label{token_eval}
Standard benchmarks often evaluate sentence encoders through sequence-level tasks, which can mask weaknesses in token representations~\citep{conneau2018xnli, xtreme, DBLP:conf/iclr/EnevoldsenCKKMS25}. Following~\citet{sonar2}, we instead focus on tasks that isolate the embeddings of individual tokens, providing a more rigorous assessment of cross-lingual token quality. For word alignment, we test whether tokens across languages align semantically. For sequence tagging, we feed the classifier head only the embedding of a single word to directly probe the quality of its token-level representations across languages.

\paragraph{Word alignment} Given a translation pair $x, y$, the task of word alignment identifies which words correspond to each other semantically. More formally, given a source–target sentence pair $x$, $y$ with word lengths $w_x$, $w_y$, respectively, we infer a binary matrix $M\in\{0,1\}^{w_s \times w_y}$, where $M_{ij}=1$ represents that the $i$-th word in the source sentence aligns semantically with the $j$-th word in the target sentence. We present an example in Table~\ref{example_datasets}.

Many current word alignment methods derive alignments based on the similarity of token embeddings~\citep{simalign, DBLP:conf/acl/AzadiFD23}. We define the \textit{token similarity matrix} $S\in \mathbb{R}^{n\times m}$; $S_{i,j}:=sim(\mathcal{E}(t_x[i]), \mathcal{E}(t_y[j]))$, where \textit{sim} is a similarity measure (we use cosine similarity), $t_x[i]$ and $t_y[j] \in V$ are the $i$-th and $j$-th tokens of the source and target sentences $x$ and $y$. An \textit{extraction} method is then applied to convert these similarities into a discrete alignment. We use Itermax~\citep{simalign} to extract alignments and report Alignment Error Rate (AER)~\citep{DBLP:conf/acl/OchN00}. The full list of alignment datasets is in Appendix~\ref{alignment_datasets}.




\paragraph{Sequence tagging}
We evaluate token-level cross-lingual representations using standard sequence labelling tasks, where each token $x_i$ in an input sequence $x = (x_1, \dots, x_n)$ is assigned a discrete label $\{1, \dots, N\}$. To isolate the contribution of the encoder, we train a linear classification head $C \in \mathbb{R}^{d \times N}$ placed on top of the token embeddings while keeping the encoder parameters fixed. The classifier is trained only on English data and subsequently evaluated in other languages (zero-shot); see Appendix~\ref{seq:implement} for implementation details. 

We consider a diverse set of benchmarks, including Massive (Slot Filling)~\citep{fitzgerald2022massive}, PAN-X (Named Entity Recognition)~\citep{DBLP:conf/acl/PanZMNKJ17}, UPOS (Part-of-Speech Tagging)~\citep{DBLP:conf/lrec/NivreMGHMPSTZ20}, and WiC (Word Sense Disambiguation)~\citep{DBLP:conf/naacl/PilehvarC19}. WiC is not technically a sequence tagging task; we adapt it by extracting the contextual embeddings of the target word in each sentence, concatenating the resulting vectors, and feeding them into a classifier $C \in \mathbb{R}^{2d \times 2}$. When a target word is split into multiple subword tokens, we represent it by averaging the corresponding embeddings. Further dataset details and links are provided in Appendix~\ref{seq_tagging_datasets}. 

Finally, to handle annotation–tokeniser mismatches, we assign labels by maximal character overlap, breaking ties by selecting the first segment.

%% file: latex/5_results.tex
\begin{table*}[t]
\centering
\setlength{\tabcolsep}{3pt}
\resizebox{0.8\textwidth}{!}{%
\begin{tabular}{lccccc}
\toprule
                & \textbf{AER ($\downarrow$)} & \textbf{PAN-X ($\uparrow$)}          & \textbf{Massive ($\uparrow$)}        & \textbf{UPOS ($\uparrow$)}          & \textbf{WiC ($\uparrow$)}\\
\midrule
\textit{Pre-trained encoders} \\
XLM-R     & 0.307                                           & 0.551          & 0.220          & 0.527          & 0.538          \\
XLM-Align       & 0.187                                                & 0.532          & 0.187          & 0.522          & 0.556          \\
LaBSE           & 0.232                                      & 0.589          & 0.431          & 0.551          & 0.538          \\
MEXMA           & \textbf{0.171}                            & 0.593          & 0.396          & 0.553          & 0.565          \\
SONAR           & 0.183                                         & 0.582          & 0.402          & 0.548          & 0.576          \\

\midrule
\textit{SONAR + additional objectives}
\\
SONAR Loss~\tiny{\citep{sonar}}     & 0.179                                             & 0.588          & 0.426          & 0.554          & 0.581          \\
SO Loss~{\tiny\citep{DBLP:conf/eacl/DouN21}}         & 0.178                                         & 0.597          & 0.410          & \textbf{0.577} & 0.574          \\

WordOT~{\tiny\citep{alqahtani-etal-2021-using-optimal}}  & 0.320 &      0.609    &     0.417      & 0.561  & \textbf{0.585}          \\

WACSE~{\tiny\citep{miao-etal-2024-enhancing}}  & 0.205                                         & 0.600          & 0.409          & 0.569 & 0.573          \\

OmniSONAR-Token~\tiny{\citep{sonar2}} & 0.180                                           & 0.610          & 0.417          & 0.574    & 0.583          \\
SALT \salt{0.05} (ours)          & \textbf{0.171}                            & \textbf{0.622} & \textbf{0.452} & 0.571          & \textbf{0.585} \\



\bottomrule
\end{tabular}}
\caption{Cross-lingual token-level evaluation across five benchmarks for frozen pre-trained encoders and SONAR trained with additional objectives. SALT achieves the best or joint-best performance on four of five benchmarks.}
\label{main-results}
\vspace{-5pt}
\end{table*}

\section{Results}
Table~\ref{main-results} compares SALT with pre-trained encoders and alternative fine-tuning objectives for SONAR across five cross-lingual token benchmarks. Among the pre-trained models, MEXMA achieves the lowest AER (0.171), while SONAR remains competitive on the other tasks. Continued training with the SONAR loss improves all token-level results, suggesting that SONAR was primarily optimised for sentence mining rather than token representations. We also compare SALT with the dedicated Awesome-Align word aligner~\citep{DBLP:conf/eacl/DouN21} (see Appendix~\ref{awesome-ablation-section}).

Among the fine-tuning objectives, SALT outperforms in word alignment (AER), PAN-X, Massive, and WiC, showing that span-level objectives provide complementary signals beyond sentence- or token-level training. Gains are particularly pronounced on tasks requiring rich cross-lingual semantics (PAN-X, Massive) and on word alignment. On UPOS, SALT achieves a score of 0.571 (vs. 0.548 SONAR), but is slightly outperformed by SO Loss (0.577). This may be due to UPOS being language-specific, with limited cross-lingual transfer relative to other tasks (Appendix~\ref{seq_tagging_UPOS})


Taken together, the results show that SALT achieves the best overall performance among the considered strategies, and suggest that incorporating span-level supervision is an effective and principled way to enrich multilingual sentence representations beyond sentence-level objectives alone.

Does improved token-level representation come at the cost of sentence-level quality? For cross-lingual sentence mining, SALT consistently improves over SONAR on xsim and xsim++ (Table~\ref{tab:merged-eval}). For MTEB classification tasks, SALT matches SONAR on XNLI and MIntent and improves on STS17. Overall, SALT does not sacrifice sentence-level quality and, in several cases, improves it.



\begin{table}[t]
\centering
\setlength{\tabcolsep}{4pt}
\small
\begin{tabular}{lccccc}
\toprule
& \multicolumn{2}{c}{\textbf{Mining ($\downarrow$) }} & \multicolumn{3}{c}{\textbf{Classification ($\uparrow$)}} \\
\cmidrule(lr){2-3} \cmidrule(lr){4-6}
& \textbf{xsim} & \textbf{xsim++} 
& \textbf{XNLI}  & \textbf{STS17} & \textbf{MIntent} \\
\midrule
\small SONAR & \small 0.2 & \small 9.9 & \small 0.61 & \small 0.65 & \small 0.58 \\
\small SALT  & \small \textbf{0.1} & \small \textbf{8.8} & \small 0.61 & \small \textbf{0.66} & \small 0.58 \\

\bottomrule
\end{tabular}
\caption{Cross-lingual sentence mining (FLORES-80) and English-only MTEB classification performance.}
\label{tab:merged-eval}
\vspace{-10pt}
\end{table}

\begin{figure}
  \centering
\includegraphics[trim={0cm 0cm 0cm 0cm},clip, width=\columnwidth]{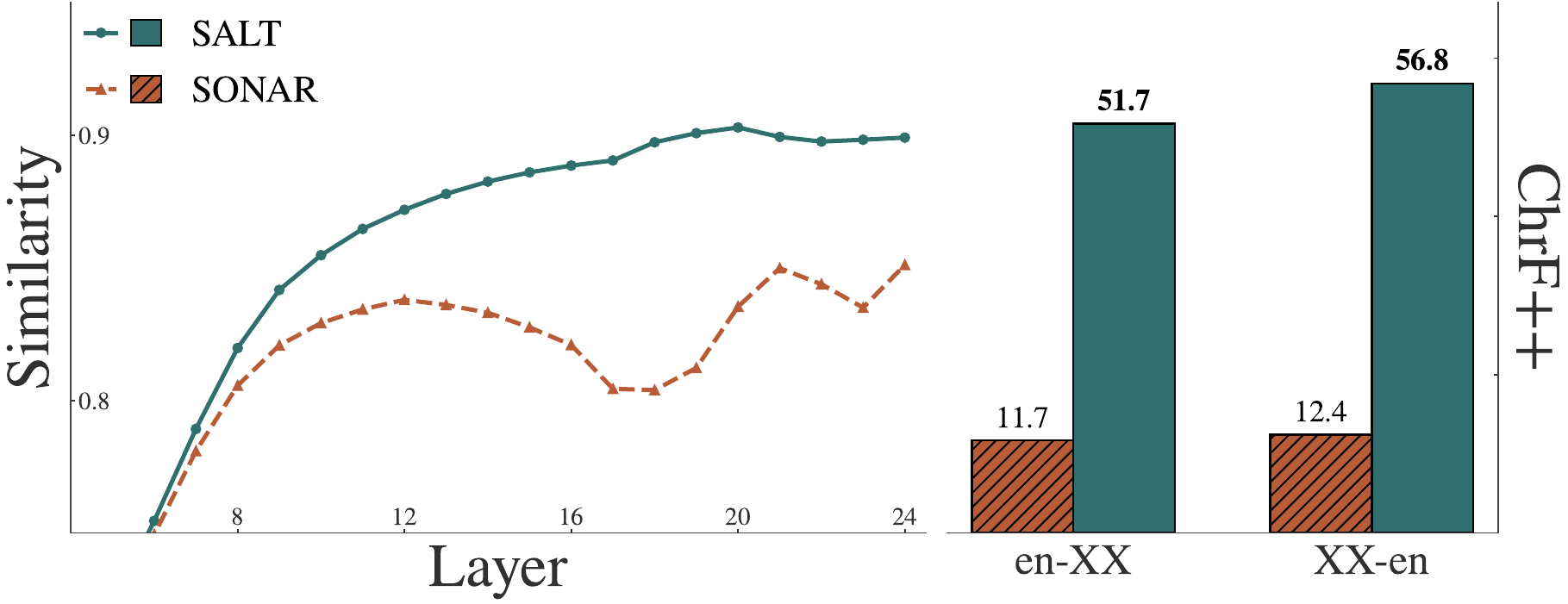}
    \caption{Left: cosine similarity of span pairs across layers. SALT shows a better similarity than SONAR across all layers. Right: translation performance for span embeddings. Both experiments use CrossSpan.}
    \label{span_embeddings}
    \vspace{-10pt}
\end{figure}

\subsection{Analysis}
\label{analysis_sec}

\paragraph{Span embeddings} 
SALT supports span-level representations by averaging token embeddings within each span. We find that these span representations are more semantically informative than their SONAR counterparts. On the CrossSpan test set of high-quality aligned cross-lingual spans, SALT representations of aligned spans consistently exhibit higher cosine similarity across layers (Figure~\ref{span_embeddings}, left), indicating stronger cross-lingual alignment.

A by-product of our span-translation objective is that the learned representations can be directly translated with a \textbf{frozen} SONAR decoder (Figure~\ref{span_embeddings}, right). SALT span embeddings yield high-quality translations, indicating that they preserve rich semantic information for downstream generation. 

\paragraph{Intermediate representations} 
Sentence encoders typically exhibit a \textit{parabolic} behaviour, where intermediate layers achieve the strongest cross-lingual alignment~\citep{simalign}. This arises because cross-lingual token representations emerge in mid-layers but are not necessarily preserved in the final layers, which are more specialised for the pretraining objective. The SALT loss shifts this pattern, with deeper layers becoming more cross-lingually aligned for both spans (Figure~\ref{span_embeddings}, left) and words (Appendix~\ref{word_embeddings_alignment}).

\paragraph{Geometry of the embedding space} We compare the geometry of word embeddings produced by SALT, SONAR, and MEXMA. Table~\ref{geometry_tasks} summarises the results across all four analyses: word retrieval, language separability, isotropy, and hubness (see Appendix~\ref{geometry_tasks_exp} for more details on experiments). 

We merge the word alignment datasets to construct a word retrieval task. For each English word, we retrieve its nearest neighbours by cosine similarity and evaluate against gold-aligned targets. SALT achieves the highest recall.

A good cross-lingual representation should minimise language-specific structure. We train a linear classifier to predict language ID from word embeddings; SALT embeddings are significantly harder to classify, indicating more language-invariant representations than SONAR and MEXMA.

Isotropy captures whether the embedding space is dominated by privileged directions, leading to uneven use of representational capacity. We measure it via mean cosine similarity over random word pairs, and via uniformity~\citep{DBLP:conf/icml/0001I20}. MEXMA is highly anisotropic. SONAR and SALT produce more isotropic spaces by these measures, though as we show below, SONAR's apparent isotropy masks a severe hubness problem.

Hubness measures the tendency of a few embeddings to dominate nearest-neighbour retrieval regardless of semantics~\citep{DBLP:journals/jmlr/RadovanovicNI10}. We quantify it via the skewness of the $k$-occurrence distribution, which counts how often an embedding appears in another word's nearest neighbours. Zero skewness indicates a uniform space, while high skewness reflects hub collapse. SONAR exhibits severe hubness (11.5 skewness), while MEXMA and SALT are substantially better behaved. 

Taken together, these results reveal contrasting geometric failures in the baselines. MEXMA produces an anisotropic, language-stratified space, while SONAR suffers from severe hub collapse. SALT avoids both, yielding a better-distributed cross-lingual token space (see Appendix~\ref{plot_pca} for a qualitative visualisation). 

\begin{table}[t]
\centering
\setlength{\tabcolsep}{3pt}
\footnotesize
\begin{tabular}{lcccc}
\toprule
 & \small \textbf{Word Ret. $\uparrow$} & \small \textbf{Lang. ID $\downarrow$} & \small \textbf{Isotropy  $\downarrow$} & \small \textbf{Hub $\downarrow$} \\
\midrule
 &  &  & sim / unif &  \\
MEXMA & 63\% & 0.49 & 0.44 / -2.2 & 4.0 \\
SONAR & 62\% & 0.46 & 0.04 / \textbf{-3.8} & 11.5 \\
SALT  & \textbf{65\%} & \textbf{0.35} & \textbf{0.03} / -3.6 & \textbf{3.5} \\
\bottomrule
\end{tabular}
\caption{Word retrieval (Recall@5), language identification (F1), isotropy (mean cosine similarity and uniformity) and hubness (skewness) for three encoders.}
\label{geometry_tasks}
\vspace{-10pt}
\end{table}

\begin{table}[t]
\resizebox{1\columnwidth}{!}{%
\begin{tabular}{lrrrr}
\toprule
                & \multicolumn{2}{c}{MEXMA}                              & \multicolumn{2}{c}{XLM-Align}                        \\ 
      & \multicolumn{1}{c}{AER}   & \multicolumn{1}{c}{SeqTag} & \multicolumn{1}{c}{AER} & \multicolumn{1}{c}{SeqTag} \\ 
      \midrule
Frozen          & \multicolumn{1}{c}{0.171} & 0.527                      & 0.187                   & 0.449                      \\
SO Loss         & \textbf{0.168}                     & 0.537                      & 0.181                   & 0.484                      \\
WordOT          & 0.17                      & 0.530                      & 0.183                   & 0.460                      \\
OmniSONAR-Token & \textbf{0.168}                     & 0.551                      & 0.181                   & 0.470                      \\
SALT (ours)     & \textbf{0.168}                     & \textbf{0.555} & \textbf{0.179}                   & \textbf{0.493}     \\
\bottomrule
\end{tabular}}

\caption{Experiments fine-tuning additional encoders. We average the sequence tagging tasks (SeqTag).}
\label{additional_encoders}
\end{table}

\subsection{Ablations}

\paragraph{Generalisation to other encoders} We apply SALT fine-tuning to MEXMA and XLM-Align to assess whether its benefits extend beyond SONAR. Table~\ref{additional_encoders} summarises the results (using an unfrozen SONAR decoder, see Appendix~\ref{extra_encoders}). SALT achieves the strongest overall performance with both encoders, demonstrating that it improves performance across different multilingual encoders.

\paragraph{CASE span extraction} Do SALT’s improvements stem from an additional external signal, given that LLaMA-70B was used for span extraction? To investigate this, we compare results when training SALT with the self-supervised span extraction heuristic CASE. Results are shown in Table~\ref{llmvslocal}. CASE yields better performance on Massive, whereas supervision from LLaMA-70B is superior on PAN-X, UPOS, and WiC. Nevertheless, the difference is marginal: CASE would outperform all the baseline losses and encoders in Table~\ref{main-results} except for UPOS. This suggests that the training regime is primarily responsible for SALT's success.

\begin{table}[t]
\small
\setlength{\tabcolsep}{3pt}
\resizebox{1\columnwidth}{!}{%
\begin{tabular}{lccccc}
\toprule
                & \textbf{AER} & \textbf{PAN-X}          & \textbf{Massive}        & \textbf{UPOS}          & \textbf{WiC}\\
\midrule
SALT (LLaMA) & 0.171          & \textbf{0.622}            & 0.452                       & \textbf{0.571}            & \textbf{0.585}          \\
SALT (CASE)      & 0.171 & 0.613                     & \textbf{0.459}              & 0.568                     & 0.584                 \\
\bottomrule
\end{tabular}}
\caption{Cross-lingual token-level evaluation for SALT using LLaMA-70B spans or the self-supervised CASE.}
\label{llmvslocal}
\end{table}

\begin{table}[t]
\setlength{\tabcolsep}{1pt}
\resizebox{1\columnwidth}{!}{%
\begin{tabular}{lcc}
\toprule
                                                                                                 & \small \textbf{AER $\downarrow$}& \small \textbf{SeqTag $\uparrow$}
                                                                                                \\
\midrule
\small SONAR (pre-trained)                                                                                          & \small 0.183        & \small 0.527                   \\
\small $\mathcal{L}_{CON_{Span}}$                                                 &   \small 0.178    &    \small 0.538              \\
\small $\mathcal{L}_{MT_{Span}}$                                                        &   \small   0.186                            & \small    0.522                                     \\
\small $\mathcal{L}_{INT}$                                                              & \small  0.174                               &  \small 0.551                                        \\
\small $\mathcal{L}_{CON_{Span}}    + \mathcal{L}_{MT_{Span}}$ &   \small   0.179                            &   \small  0.551                                     \\
\small $\mathcal{L}_{MT_{Span}} + \mathcal{L}_{INT}$ 

&   \small  0.177                             &   \small  \underline{0.554}                                     \\
\small $\mathcal{L}_{CON_{Span}} + \mathcal{L}_{INT}$         &  \small  \textbf{0.170}                              &  \small      0.552                                  \\
\small
$\mathcal{L}_{CON_{Span}}    + \mathcal{L}_{MT_{Span}} +  \mathcal{L}_{INT}$  (SALT)                                                                                      &   \small  \underline{0.171}                             &       \small \textbf{0.557}           \\                       \bottomrule
\end{tabular}}
\caption {Fine-tuning SONAR with different losses. We average the sequence tagging tasks (SeqTag).}
\label{ablation}
\vspace{-10pt}
\end{table}

\paragraph{Additional downstream tasks} We evaluate SALT on two additional downstream tasks: hallucination detection and label projection. SALT achieves the best average performance on label projection and the strongest overall results on hallucination detection (see Appendix~\ref{downstream}).

\paragraph{Impact of the different losses}


Table~\ref{ablation} ablates each SALT loss. Post-training on $\mathcal{L}_{{CON}_{Span}}$ or $\mathcal{L}_{INT}$ improves both tasks, with the latter being the strongest single loss. $\mathcal{L}_{{MT}_{Span}}$ degrades the pre-trained model and worsens word alignment when combined with either term. However, it improves sequence tagging in combination with the other losses (e.g., $\mathcal{L}_{{CON}_{Span}}$: 0.538 $\rightarrow$ 0.551 when adding $\mathcal{L}_{{MT}_{Span}}$), and the full SALT objective achieves the best SeqTag score (0.557). $\mathcal{L}_{{CON}_{Span}}+\mathcal{L}_{INT}$ achieves a marginally better AER (0.170 vs.\ 0.171). No configuration is best on both metrics, so we use the full objective, which gives the best sequence tagging and near-best alignment. 

%% file: latex/6_conclusions.tex
\section{Conclusions}
We introduce SALT, a lightweight post-training method that improves cross-lingual token representations through span-aligned supervision. SALT outperforms competitive encoders and fine-tuning strategies on token-level benchmarks while preserving sentence-level performance, suggesting that span-level supervision can be a powerful inductive bias for multilingual representation learning. 


%% file: latex/99_ethics.tex
\section*{Limitations}
Despite the strong empirical performance of SALT, several limitations remain.

SALT relies on parallel corpora to extract aligned spans, which may limit its applicability to truly low-resource languages. In addition, the self-supervised CASE heuristic assumes that the underlying encoder already provides reasonably good cross-lingual token representations, making adaptation to unseen or severely underrepresented languages more challenging; in those cases, LLM-based spans should be used instead.

Our method also uses average pooling to construct span embeddings. While lightweight and architecture-agnostic, this may fail to capture finer internal structure within longer or compositionally complex spans.

Finally, we only evaluate SALT as a post-training intervention. Exploring span-level supervision during large-scale pre-training could potentially yield larger improvements, but was beyond the computational scope of this work.

\section*{Ethical Considerations}
This work aims to improve multilingual language representations for token-level cross-lingual tasks, which may help broaden language accessibility and support underrepresented languages in NLP. However, multilingual representation models may also be used in sensitive applications such as surveillance or profiling, and we therefore encourage responsible deployment and evaluation.

Our training data are derived from publicly available parallel corpora from the NLLB Primary dataset, which may contain societal or cultural biases present in web-scale multilingual text. As a result, SALT may inherit biases from the training data or pretrained models.

We also conducted a human evaluation of span quality using native speakers. Participation was voluntary, annotators were informed about the study and their right to withdraw, and no personally identifiable information was collected beyond anonymised annotator IDs. The study received ethics approval from our institution.

%% file: latex/98_appendix.tex
\newpage
\section{A note on segmentation}
We define words as whitespace-delimited units of text. For languages that do not use whitespace segmentation—namely scripts \textit{Hans, Hant, Jpan, Thai, Laoo, Khmr, Mymr, Tibt} following the SONAR notation—we treat tokeniser tokens as words. When computing word embeddings, we obtain them by averaging the embeddings of the tokens that compose each word. Throughout the paper, we use the FLORES-200 (NLLB) language identifiers (e.g., \texttt{zho\_Hans})~\citep{nllb}.

\section{Spans extraction}
\subsection{Extracting spans with an LLM}
\label{span:extraction}
For the LLM-based method, we use LLaMA-70B Instruct, quantised to 8-bit precision using Bitsandbytes. The prompt used for span extraction is provided in Figure~\ref{prompt:span_llm}.

The generated spans are subsequently mapped back to SONAR tokens. In the vast majority of cases, this mapping is unambiguous; when a SONAR token is split across multiple spans, we assign it to the span with the greatest string overlap.

\subsection{Statistics on extracted spans}
Table~\ref{table-span:stat} reports summary statistics of the extracted spans. We observe that span lengths vary across languages, with CASE tending to extract longer spans in both source and target sides. In particular, Chinese and Hindi exhibit relatively longer spans compared to other languages, often involving a larger number of target tokens. Using spaCy’s \texttt{en\_core\_web\_sm} model to analyse POS tags on the English side, we find that nouns are most frequent, followed by adpositions. We also observe variation in POS tag distributions across languages and span sources; further details are provided in Table~\ref{statistics_span:table}.

For CASE, we used the Benepar parser, which achieves 95.17 F1 on the English WSJ test set\footnote{See \href{ https://github.com/mrdrozdov/self-attentive-parser-with-extra-features}{ https://github.com/mrdrozdov/self-attentive-parser-with-extra-features}}, making parser errors relatively infrequent.

\section{Experiment details}
\label{exp:details}

\paragraph{Pre-processing} We apply exactly the same pre-processing and normalisation as SONAR, namely the script \href{https://github.com/facebookresearch/stopes/blob/main/stopes/pipelines/monolingual/utils/text_normalizer.py}{https://github.com/facebookresearch/stopes/blob/main/stopes/pipelines/monolingual/utils/text\_normalizer.py}

\subsection{Languages studied}
\label{langs_info}
Table~\ref{all_langs} shows the language splits used for training and evaluation. Training languages are drawn from the NLLB Primary dataset, while the test languages correspond to those covered by our cross-lingual token-level benchmark, including Word Alignment, PAN-X, MASSIVE, UPOS, and WiC. For sentence retrieval, we use the 57 test languages; we only evaluate the MTEB tasks on English. 

\subsection{Alignment datasets}
\label{alignment_datasets}
We evaluate a total of 18 language pairs. As gold-standard alignments, we use the XL-WA dataset~\citep{xlwa} for 10 language pairs; we were unable to obtain the four additional subsets. For en-hi, en-fr, en-fa, and en-cs, we adopt the same datasets as in~\citet{simalign}. For en-zh, en-ro, and en-de, we use the datasets introduced by~\citet{DBLP:conf/eacl/DouN21}, and for en-ja we use the KFTT dataset~\citep{neubig11kftt}.
\subsection{Sequence tagging datasets}
\label{seq_tagging_datasets}

We use UPOS and PAN-X from XTREME~\citep{xtreme}, following its standard dataset configurations. UPOS covers 51 languages using Universal Dependencies treebanks for part-of-speech tagging, while PAN-X covers 40 languages for named entity recognition. The language and treebank details, along with the dataset splits, are documented in the \href{https://huggingface.co/datasets/google/xtreme}{XTREME repository}.

For word-in-context evaluation, we use MCL-WiC, which covers 6 languages and is available from its \href{https://github.com/zhestyatsky/MCL-WiC/tree/main/data/MCL-WiC}{online repository}. For slot filling, we use MASSIVE, which covers 52 languages and is available from the \href{https://huggingface.co/datasets/AmazonScience/massive}{MASSIVE repository}.

\subsection{Sequence tagging evaluation}
\label{seq:implement}
We use up to 10{,}000 training examples and 5{,}000 test examples per language, doubling these limits for WiC, while filtering out sequences exceeding 514 tokens. A single linear classifier (\texttt{Linear(dim, n\_classes)}) is trained on frozen encoder embeddings using Adam with a learning rate of \(10^{-3}\), cross-entropy loss, and a batch size of 4096. We train with early stopping (patience \(=10\)) based on a 90/10 train-validation split. All embeddings are L2-normalised before classification.

\subsection{Losses implementations details}
\label{losses:implement}
We briefly describe each loss for completeness, while referring readers to the original papers for detailed formulations and motivation. We focus on the implementation details and hyperparameters used in our setup.

\paragraph{SO Loss} The Self-Objective loss~\citep{DBLP:conf/eacl/DouN21} is defined as
\begin{equation}
\begin{aligned}
\mathcal{L}_{\mathrm{SO}}
= \sum_{i,j} M_{ij} \frac{1}{2}
&\left(
\frac{\operatorname{softmax}(S_{i,*})}{n} \right.\\
&\left. + \frac{\operatorname{softmax}(S_{*,j})}{m}
\right)
\end{aligned}
\end{equation}
where $M_{ij} \in \{0,1\}$ is the alignment obtained with the Argmax method, $S$ is the token similarity matrix between source and target tokens. We follow~\citet{sonar2} and introduce a temperature hyperparameter $\tau=500$ at the softmax. We use Argmax~\citep{simalign} for word alignment extraction. 

\paragraph{OmniSONAR-Token} \citet{sonar2} use the SO Loss with the interpolation objective as defined in Equation~\ref{interpolation_loss}

\begin{equation}
 \mathcal{L_\mathrm{OmniSONAR-Token}} =  \mathcal{L}_{INT} +  \mathcal{L}_{SO}
\label{eq:so_loss}
\end{equation}

\paragraph{WordOT} WordOT~\citep{alqahtani-etal-2021-using-optimal} uses Optimal Transport to
learn a soft alignment between source and target tokens. The objective is
defined as
\begin{equation}
\begin{aligned}
\mathcal{L}_{\mathrm{WordOT}}
= -S_{\epsilon}(\mu_x,\mu_y)
+ \lambda \sum_{i=1}^{|y|}
\|H_y^i-H_{y,0}^i\|_2^2,
\end{aligned}
\end{equation}
where $S_{\epsilon}$ is the Sinkhorn divergence between the source and
target token representations, and $H_y^i$ and $H_{y,0}^i$ denote the
fine-tuned and initial representations of the $i$-th target token,
respectively. We use $\epsilon=0.05$ and $\lambda=1$ following
\citet{alqahtani-etal-2021-using-optimal}.

\paragraph{WACSE} WACSE~\citep{miao-etal-2024-enhancing}  combines an Aligned Word Prediction task (AWP) that uses a language head, a Word Translation Ranking task equivalent to the SO Loss, and a Translation Ranking (TR) task which operates on batches of parallel sentence pairs and applies a softmax over the diagonal of the sentence similarity matrix, encouraging each source sentence to be closer to its corresponding translation than to other sentences in the batch.

\begin{equation}
    \mathcal{L}_{\mathrm{WACSE}} =  \alpha\mathcal{L}_{\mathrm{TR}} +  \beta\mathcal{L}_{\mathrm{AWP}} +  \gamma \mathcal{L}_{\mathrm{SO}}
\end{equation}

We use Argmax~\citep{simalign} for word alignment extraction (SO, AWP losses). For SO loss, we use $\tau=1$. For AWP, we only use one mask per sentence pair. We use the token embedding (encoding) matrix as the MLM head, and use a cross-entropy loss. Finally, following \citet{miao-etal-2024-enhancing}, we use $\alpha=0.8$, $\beta=0.1$, $\gamma=0.1$. 

\paragraph{SALT} We use $\tau=5.0$ in our contrastive span loss. 

\subsection{Data sampling during training}
Training uses dynamically constructed multilingual mini-batches under a fixed token budget. At each step, sentence pairs are sampled from source–target language distributions using temperature-smoothed probabilities over language pairs. Each pair is tokenised with the corresponding language-specific encoder, and any example exceeding the maximum sequence length (512) on either side is discarded.

A buffered packing strategy is then used to improve batch efficiency. Examples are first ordered by sequence cost, defined as $\max(|x|, |y|)$, where $|x|$ and $|y|$ denote the number of tokens in the source and target sentences. This reduces padding and increases the proportion of useful tokens per update.

Optimisation follows the setup described in the main paper, using AdamW with a linear warm-up over the first 1,000 steps and cosine annealing thereafter, decaying the learning rate smoothly to zero over the rest of training.

\section{Other results}
\subsection{Training dynamics}
\begin{figure*}
  \centering
\includegraphics[trim={0cm 0cm 0cm 0cm},clip, width=\linewidth]{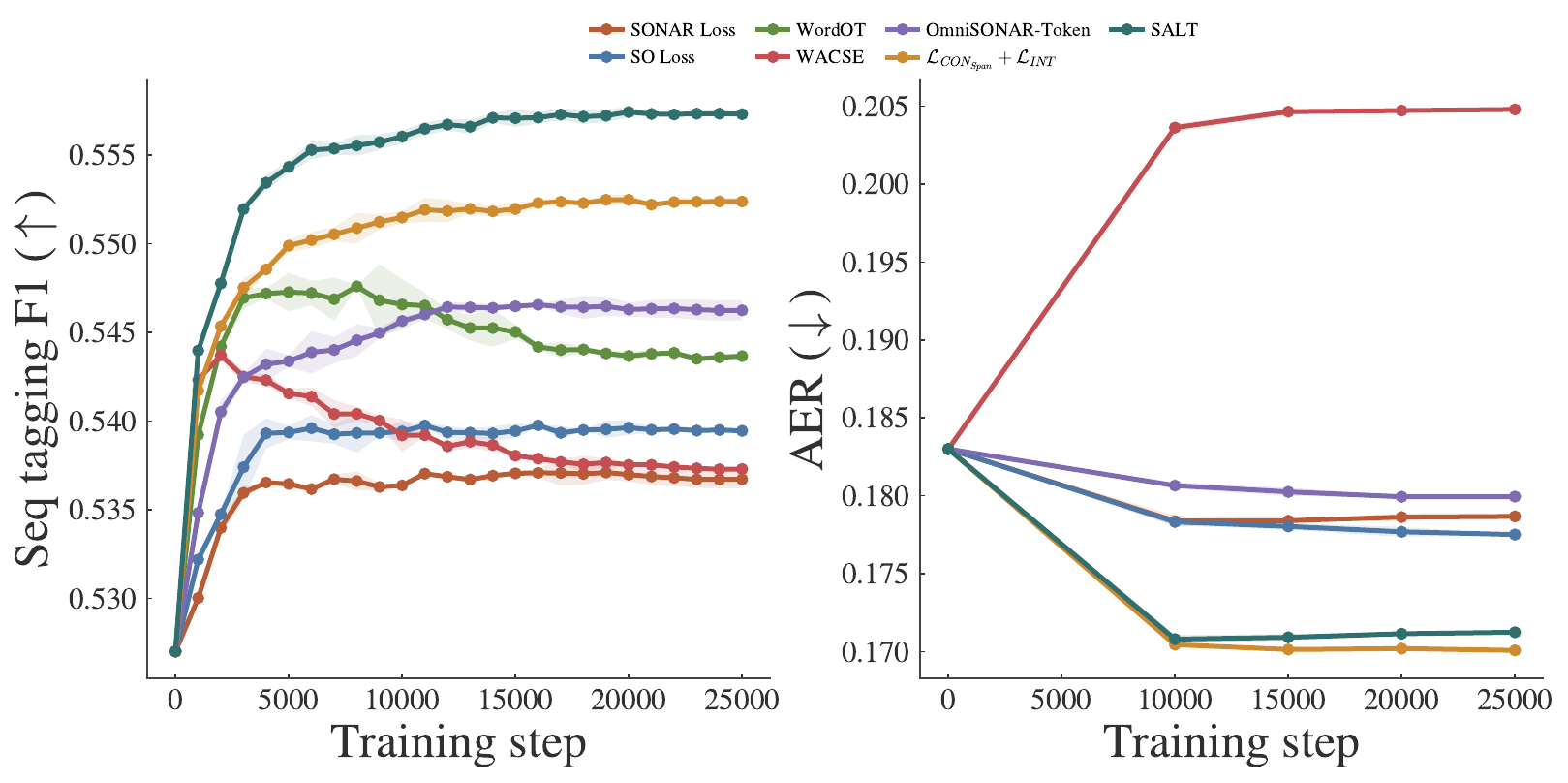}
    \caption{Sequence tagging and word alignment performance for different training losses. In the word alignment plot, we omit WordOT to improve visualisation (WordOT is poorly aligned).}
    \label{training-epochs}
\end{figure*}

We show the training dynamics in terms of sequence tagging and Word Alignment performance of the different losses studied in Figure~\ref{training-epochs}. 

\subsection{Sequence tagging results}
In this section, we provide an expanded analysis of the sequence tagging results.
\label{seq_tagging_UPOS}

We evaluate sequence tagging under three different training settings:
\begin{itemize}
\item \textbf{English}: the classifier is trained using only English data.
\item \textbf{All}: the classifier is trained jointly on data from all languages.
\item \textbf{Single}: for each target language, the classifier is trained exclusively on data from that language.
\end{itemize}

\begin{figure*}
  \centering
\includegraphics[trim={0cm 0cm 0cm 0cm},clip, width=\linewidth]{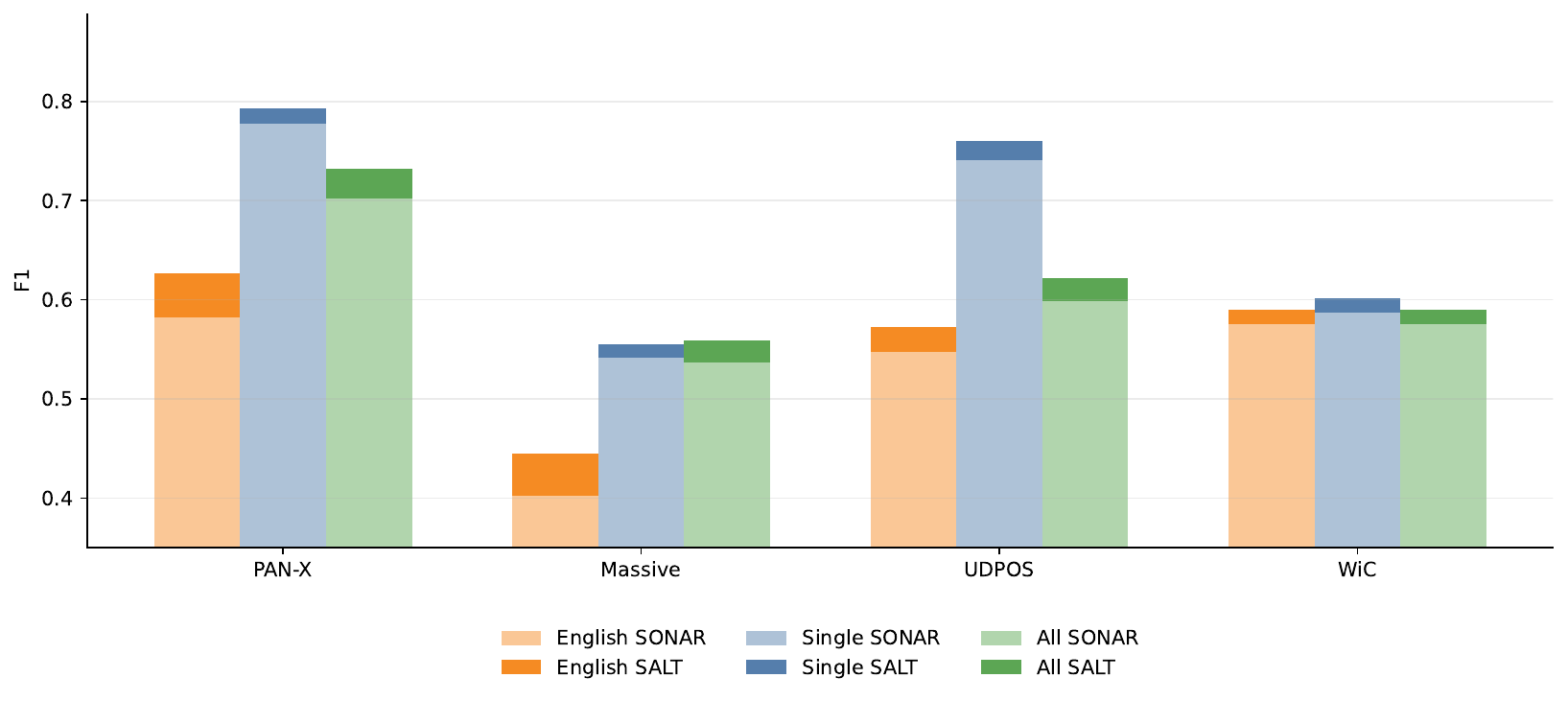}
    \caption{Performance of SALT vs SONAR on the sequence tagging tasks, on three training modes for the probe: only English data, all the data and data for every test language. SALT improves all the results. }
    \label{word-embeddings}
\end{figure*}

\begin{figure*}
  \centering
\includegraphics[trim={0cm 32cm 0cm 0cm},clip, width=\linewidth]{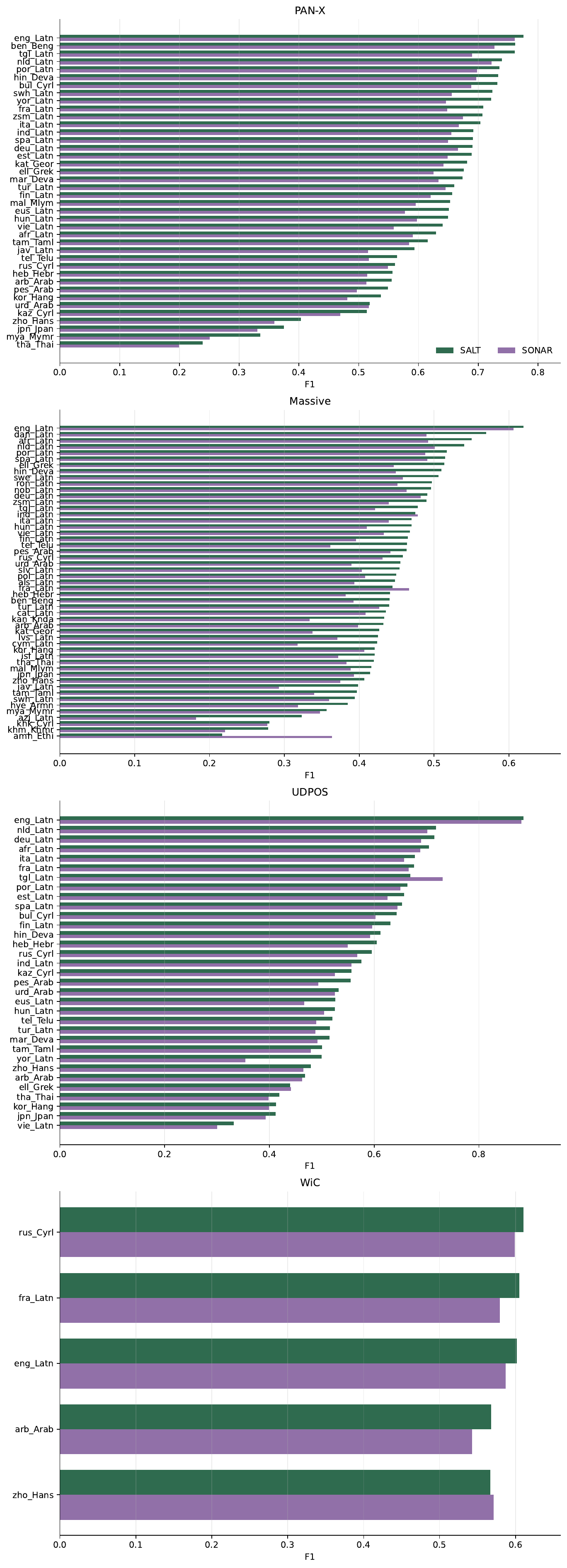}
    \caption{Performance of SALT vs SONAR on the sequence tagging tasks PAN-X and Massive, on the English setting.}
    \label{per:lang:panx}
\end{figure*}

Across all three settings, SALT consistently outperforms SONAR (Figure~\ref{word-embeddings}, Table~\ref{table:all_seq}); Figure~\ref{per:lang:panx} and \ref{per:lang:UPOS} show the performance per language on the English setting). Notably, we observe that UPOS exhibits limited cross-lingual transfer: training on a single language yields substantially better performance than multilingual training (\textbf{Single} vs.\ \textbf{All}), suggesting that the task depends strongly on language-specific characteristics.

\begin{table*}[]
\centering
\resizebox{0.8\textwidth}{!}{%
\begin{tabular}{ccccccc}
\toprule
\textbf{Task} & \textbf{SALT (single)} & \textbf{SALT (English)} & \textbf{SALT (all)} & \textbf{SONAR (single)} & \textbf{SONAR (English)} & \textbf{SONAR (all)} \\
\midrule
Massive       & 0.55                   & 0.45                    & 0.56                & 0.54                    & 0.40                    & 0.54                 \\
PAN-X         & 0.79                   & 0.63                    & 0.73                & 0.78                    & 0.58                    & 0.70                 \\
UPOS         & 0.76                   & 0.57                    & 0.62                & 0.74                    & 0.55                    & 0.60                 \\
WiC           & 0.60                   & 0.59                    & 0.59                & 0.59                    & 0.58                    & 0.58       \\         \bottomrule
\end{tabular}}
\caption{F1 scores for the sequence tagging tasks on the three training settings.}
\label{table:all_seq}
\end{table*}

\subsection{Language generalisation}
\label{test_unseen_bruto}

\begin{figure*}
  \centering
\includegraphics[trim={0cm 0cm 0cm 0cm},clip, width=\linewidth]{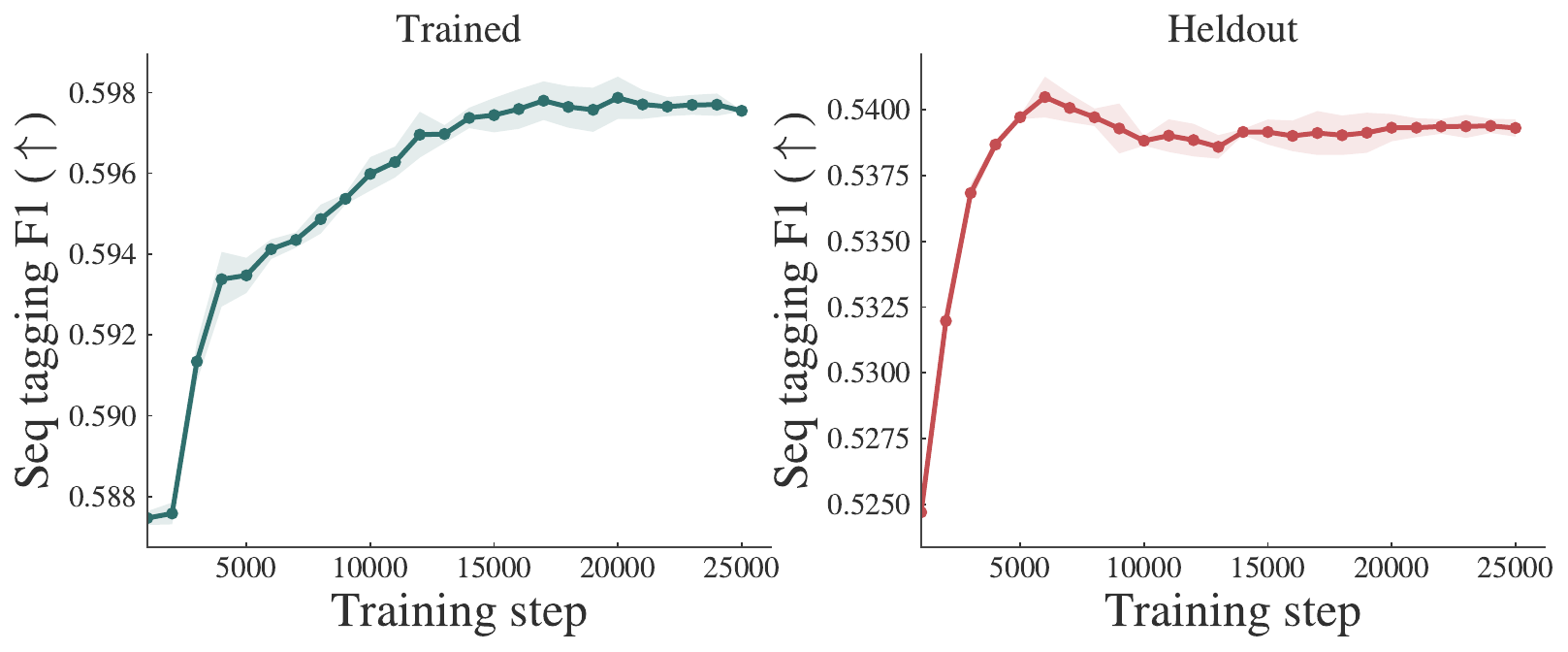}
    \caption{Sequence tagging performance throughout SALT training on trained (left) and held-out (right) languages.}
    \label{seq_tag:dynam_heldout}
\end{figure*}

We evaluate the generalisation ability of our method on held-out test languages unseen during training. Results are consistent, with SALT outperforming the other losses on these languages (see Table~\ref{test_unseen}) except on UPOS. Importantly, the training dynamics (Figure~\ref{seq_tag:dynam_heldout}) show that held-out languages benefit from SALT, particularly at earlier training stages, whereas trained languages continue to improve over a longer training horizon.

\begin{table}[]
\centering
\resizebox{1\columnwidth}{!}{%
\begin{tabular}{lccc}
\toprule
      & \textbf{PAN-X} & \textbf{Massive} & \textbf{UPOS} \\
      \midrule
SONAR Loss      & 0.562          & 0.411            & 0.530          \\
SO Loss         & 0.577          & 0.395            & \textbf{0.549} \\
WordOT          & 0.586          & 0.410            & 0.535          \\
WACSE           & 0.582          & 0.400            & 0.538          \\
OmniSONAR-Token & 0.590          & 0.402            & 0.548          \\
SALT            & \textbf{0.598} & \textbf{0.442}   & 0.544         \\ \bottomrule
\end{tabular}}
\caption{Sequence tagging performance on test languages not seen during training.}
\label{test_unseen}
\end{table}

\subsection{Comparison with Awesome-Align}
\label{awesome-ablation-section}
To compare SALT with a dedicated word alignment model, we evaluate Awesome-Align~\citep{DBLP:conf/eacl/DouN21}, which fine-tunes multilingual BERT on parallel data for word alignment. We consider both the publicly released Awesome-Align model and a model retrained using the official Awesome-Align implementation on the same 1.3M-sentence parallel corpus used for SALT, with default hyperparameters. Following~\citet{DBLP:conf/eacl/DouN21}, we use representations from the 8th layer of mBERT for word alignment.

Table~\ref{awesome-ablation} reports the results. Both Awesome-Align models improve over frozen mBERT on word alignment, PAN-X, and UPOS, but perform less well on WiC and Massive. SALT achieves the strongest performance across all five benchmarks.

\begin{table}[]

\resizebox{1\columnwidth}{!}{%
\begin{tabular}{lccccc}
\toprule
                         & \textbf{AER}       & \textbf{PAN-X}     & \textbf{Massive}   & \textbf{UPOS}     & \textbf{WiC}       \\
                         \midrule
mBERT (frozen)           & 0.202     & 0.572     & 0.132     & 0.441     & 0.569     \\
Awesome-Align (released) & 0.180     & 0.589     & 0.132     & 0.509     & 0.553     \\
Awesome-Align (ours)     & 0.201     & 0.589     & 0.139     & 0.522     & 0.556     \\
SONAR (frozen)           & 0.183     & 0.582     & 0.402     & 0.548     & 0.576     \\
SALT                     & \textbf{0.171} & \textbf{0.622} & \textbf{0.452} & \textbf{0.571} & \textbf{0.585} \\ \bottomrule
\end{tabular}}
\caption{Evaluation of Awesome-Align on our cross-lingual benchmark.}
\label{awesome-ablation}
\end{table}

\subsection{Experiments with additional downstream tasks}
\label{downstream}
We evaluate SALT, as well as other baselines, in the tasks of label projection and hallucination detection. 
\subsubsection{Label projection}
Label projection is the task of transferring sequence labels from a pivot language to a source language for which no sequence tagger is available. Given a source sentence, we first translate it into the pivot language, obtain labels for the translated sentence using a tagger trained in the pivot language, and then project these labels onto the original sentence. Following \citet{yarmohammadi-etal-2021-everything}, we perform projection by aligning the source sentence with its pivot translation using the Argmax method. We evaluate this setting on French, Russian, and Spanish using the ABSA dataset~\citep{garcia-ferrero-etal-2022-model}, and on Yoruba and Swahili using MasakhaNER~\citep{adelani-etal-2021-masakhaner}.

Table~\ref{label_proj} presents the results. SALT yields average improvements across the evaluated languages and tasks. We hypothesise that these gains stem, at least in part, from improvements in word alignment (as seen in Table~\ref{main-results}), which may lead to more accurate label projection.

\begin{table}[]
\resizebox{1\columnwidth}{!}{%
\begin{tabular}{lc}
\toprule
       & \textbf{Label Projection ($\uparrow$)} \\ \midrule
SONAR (frozen)     & 0.634                                              \\ 
SONAR Loss         & 0.634                                              \\ 
SO Loss            & 0.635                                              \\ 
SALT (ours)        & \textbf{0.636}     \\ \bottomrule                               
\end{tabular}}
\caption{Label projection results (F1). }
\label{label_proj}
\end{table}

\subsubsection{Hallucination detection for Machine Translation}

We evaluate the different losses on machine-translation hallucination and omission detection using the HalOmi dataset~\citep{dale-etal-2023-halomi}. For each machine translation pair, we compute the cosine distance between its sentence representations and assess how accurately this score ranks examples according to human-annotated error severity. Hallucinations (HAL) correspond to target content unsupported by the source, whereas omissions (OMI) correspond to source content missing from the target. 

Since a single cosine-distance score cannot distinguish between hallucinations and omissions, an example may receive a high score due to one error type even when evaluated against the other. We therefore additionally report two aggregate metrics: Severity, defined as the maximum of the hallucination and omission severity scores for each example, and Binary, indicating whether either error type is present. All metrics measure pairwise ranking accuracy.

Table~\ref{mtqe} presents the results. SALT achieves the best overall performance, particularly on the aggregated metrics, indicating that its representations are especially effective at detecting semantic mismatches caused by unsupported or missing content in either the source or target sentence.

\begin{table}[]
\resizebox{1\columnwidth}{!}{%
\begin{tabular}{lcccc}
\toprule
                & \textbf{HAL ($\uparrow$)}  & \textbf{OMI ($\uparrow$)}  & \textbf{Severity ($\uparrow$)} & \textbf{Binary ($\uparrow$)} \\ \midrule
SONAR (frozen)  & 80.6          & 81.1          & 80.6              & 80.2            \\ 
SONAR Loss      & \textbf{81.8} & 82.7          & 82.4              & 82.7            \\ 
SO Loss         & 80.1          & 80.8          & 80.8              & 80.6            \\ 
WordOT          & 60.2          & 61.8          & 61.6              & 62.8            \\ 
WACSE           & 81.7          & 79.5          & 79.6              & 80.0            \\ 
OmniSONAR-Token & 82.1          & 82.1          & 82.0              & 82.0            \\ 
SALT (ours)     & 81.3          & \textbf{83.2} & \textbf{82.7}     & \textbf{82.9} \\ 
\bottomrule

\end{tabular}}
\caption{Hallucination detection (pairwise ranking accuracy) on HalOmi using SONAR trained with different objectives. Results are macro-averaged across all 18 translation directions.}
\label{mtqe}
\end{table}

\subsection{Experiments with additional encoders}
\label{extra_encoders}
We further test the generalisation of our method by fine-tuning MEXMA and XLM-Align with SALT. Since neither encoder provides a decoder, we pair each with the pre-trained SONAR decoder for the translation objective. Because these encoders’ embedding spaces differ from SONAR’s, we leave the decoder unfrozen, which gave the best results; it is trained jointly and discarded after training. For XLM-Align (d = 768), we add a linear projection to the decoder’s input dimension (d=1024), also used only during training. We use the same training setup as for SONAR, including hyperparameters, training schedule, and random seeds, without encoder-specific tuning. Results are reported in Table~\ref{mexma_all} for MEXMA and Table~\ref{xlmalign_all} for XLM-Align. For XLM-Align, we report word alignment results from the eighth layer, following~\citet{chi-etal-2021-improving}. We expect that encoder-specific hyperparameter tuning could yield further improvements.

For MEXMA, SALT achieves the best performance across all tasks. In particular, it achieves strong results on PAN-X (0.626) and MASSIVE (0.461), while matching the strongest baselines on the remaining tasks; differences on WiC are within the noise. We observe a similar trend for XLM-Align, where SALT outperforms all baselines except on WiC. These results demonstrate that SALT can be readily applied to different encoders.

\begin{table*}
\centering
\setlength{\tabcolsep}{3pt}
\resizebox{0.8\textwidth}{!}{%
\begin{tabular}{lccccc}
\toprule
                & \textbf{AER ($\downarrow$)} & \textbf{PAN-X ($\uparrow$)}          & \textbf{Massive ($\uparrow$)}        & \textbf{UPOS ($\uparrow$)}          & \textbf{WiC ($\uparrow$)}\\
\midrule
MEXMA & \textbf{0.171}                            & 0.593          & 0.396          & 0.553          & \textbf{0.565}          \\
\midrule
\textit{MEXMA + additional objectives}
\\

SO Loss~{\tiny\citep{DBLP:conf/eacl/DouN21}}          & \textbf{0.168}              & 0.616                       & 0.398                         & 0.567                       & \textbf{0.565}            \\ 
WordOT~{\tiny\citep{alqahtani-etal-2021-using-optimal}}       & 0.170                       & 0.601                       & 0.399                         & 0.556                       & 0.564                     \\ 
OmniSONAR-Token~\tiny{\citep{sonar2}} & \textbf{0.168}              & 0.619                       & 0.453                         & 0.567                       & \textbf{0.565}            \\ 
SALT \salt{0.05} (ours)       & \textbf{0.168}              & \textbf{0.626}              & \textbf{0.461}                & \textbf{0.568}     & \textbf{0.565}  
\\

\bottomrule
\end{tabular}}
\caption{Cross-lingual token-level evaluation across five benchmarks for MEXMA with additional objectives. SALT achieves the best or joint-best performance on all benchmarks.}
\label{mexma_all}
\vspace{-5pt}
\end{table*}

\begin{table*}
\centering
\setlength{\tabcolsep}{3pt}
\resizebox{0.8\textwidth}{!}{%
\begin{tabular}{lccccc}
\toprule
                & \textbf{AER ($\downarrow$)} & \textbf{PAN-X ($\uparrow$)}          & \textbf{Massive ($\uparrow$)}        & \textbf{UPOS ($\uparrow$)}          & \textbf{WiC ($\uparrow$)}\\
\midrule
XLM-Align  & 0.187                       & 0.532                       & 0.187                         & 0.522                       & 0.556                     \\
\midrule
\textit{XLM-Align + additional objectives}
\\

SO Loss~{\tiny\citep{DBLP:conf/eacl/DouN21}}          & 0.181                       & 0.571                       & 0.232                         & \textbf{0.566}              & \textbf{0.567}            \\
WordOT~{\tiny\citep{alqahtani-etal-2021-using-optimal}}        & 0.183                       & 0.551                       & 0.213                         & 0.532                       & 0.544                     \\

OmniSONAR-Token~\tiny{\citep{sonar2}} & 0.181                       & 0.562                       & 0.225                         & 0.565                       & 0.526                     \\
SALT \salt{0.05} (ours)      & \textbf{0.179}              & \textbf{0.605}              & \textbf{0.245}                & \textbf{0.566}              & 0.554                     \\
\bottomrule
\end{tabular}}
\caption{Cross-lingual token-level evaluation across five benchmarks for XLM-Align with additional objectives. SALT achieves the best or joint-best performance on four out of five benchmarks.}
\label{xlmalign_all}
\vspace{-5pt}
\end{table*}

\section{Analysis}
\subsection{Visualisation}
\label{plot_pca}
\begin{figure*}[t]
  \centering
\includegraphics[trim={0cm 0.4cm 0cm 0cm},clip, width=1\textwidth]{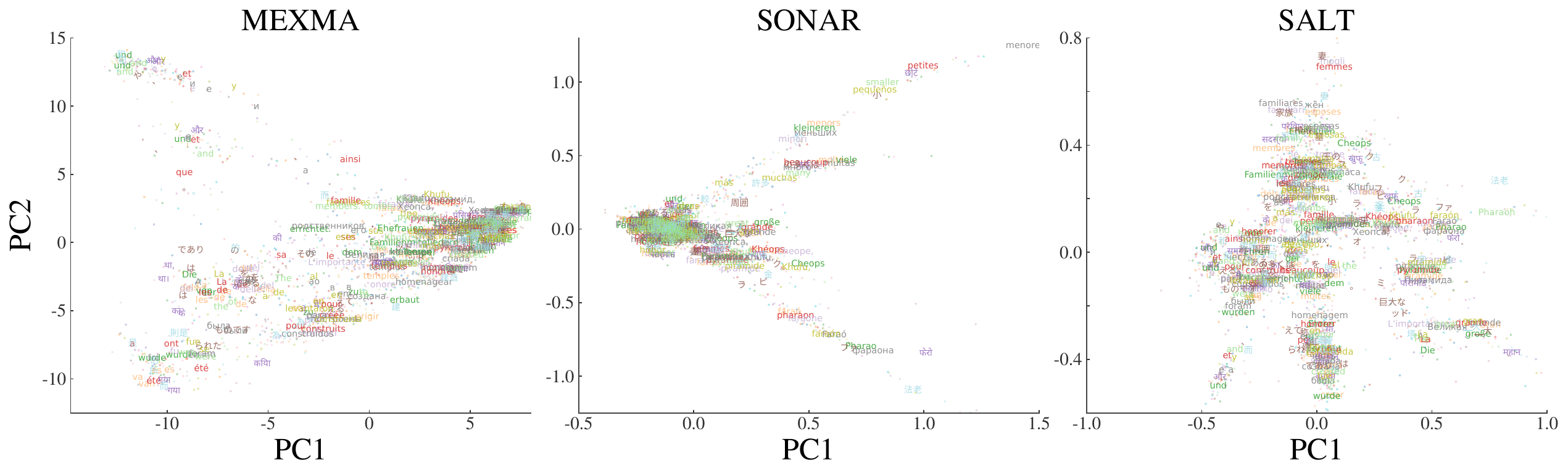}
    \caption{Principal component projections of token embeddings for the same sentence across 80 languages. Each colour corresponds to a language. SALT produces a more distributed and language-agnostic embedding space. Sentence: \textit{The great pyramid was created to honor the Pharaoh Khufu, and many of the smaller pyramids, tombs, and temples were built to honor Khufu's wives and family members.}}
    \label{pca}
    \vspace{-7pt}
\end{figure*}

To qualitatively compare the multilingual embedding spaces of SONAR and the baseline model, we visualise token embeddings for the same sentence across 80 languages using principal component analysis (PCA). Figure~\ref{pca} shows that SALT produces a more distributed and language-agnostic embedding space.

\subsection{Word embeddings alignment}
\label{word_embeddings_alignment}
\begin{figure}
  \centering
\includegraphics[trim={0cm 0cm 0cm 0cm},clip, width=\columnwidth]{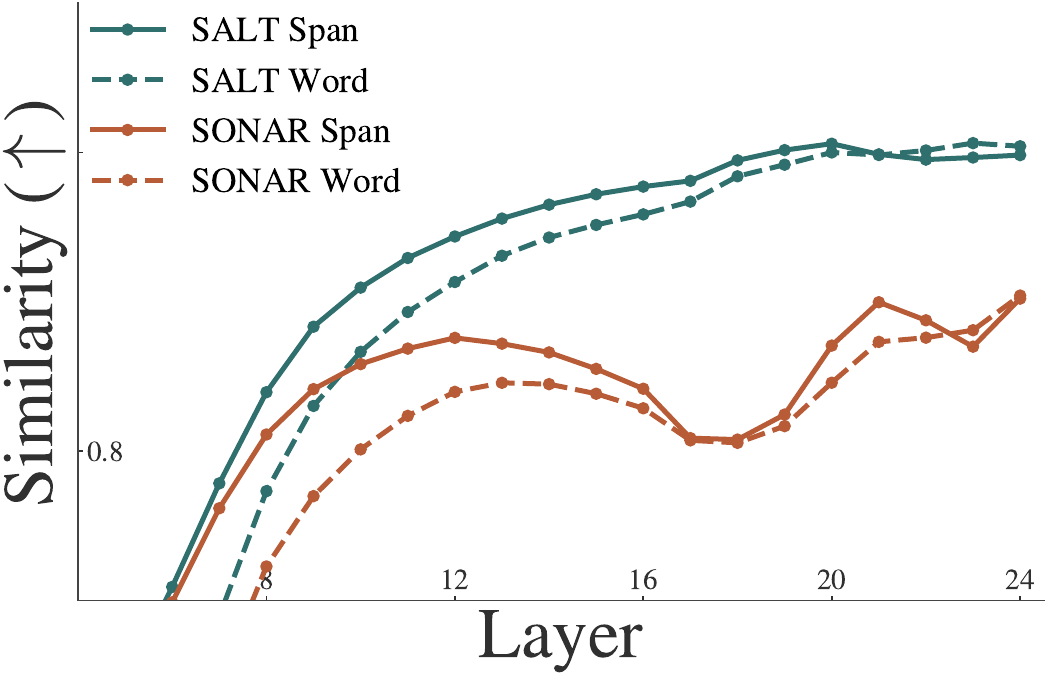}
    \caption{Word and span similarity for SONAR and SALT across layers.}
    \label{across_layers_cos}
\end{figure}

Figure~\ref{across_layers_cos} shows the cosine similarity of spans and words. We use high-quality reference spans from CrossSpan and split them into words or multi-word spans based on length. Note that this applies only to this figure; in other experiments, spans may also consist of single words.

\subsection{Geometry analysis}
\label{geometry_tasks_exp}
For word retrieval, we use our word alignment dataset (Appendix~\ref{alignment_datasets}): first, we aggregate word embeddings by averaging token embeddings (given space segmentation). Second, for every English word that has a defined target, we find the 5 nearest neighbours using cosine similarity.

We run the remainder of our experiments for geometry analysis on the FLORES-80 devtest dataset.

We evaluate language identification with a linear probe on L2-normalised FLORES word embeddings. For each model, we train a single \texttt{nn.Linear(1024, 80)} classifier to predict the language from individual word embeddings, using the first 500 FLORES sentences per language; 10\% of these sentences are held out for validation and the remaining saved sentences are used for testing. Training uses AdamW with learning rate 10e-3, batch size 8192, weight decay 0 and early stopping after 50 epochs without validation macro-F1 improvement. Duplicate training words are removed per language. 

For isotropy and hubness, we sample 10,000 random pairs of different sentences in different languages. For hubness, we use $k=10$ nearest neighbours. 

\subsection{Qualitative analysis}
We take the sentence
\begin{quote}
    \textit{Most insects have the advantage of being able to fold their wings back along the body.}
\end{quote}
 and perform word-level retrieval (top-5 nearest neighbours) across sentences in 80 languages (see Table~\ref{word:most},\ref{word:insect}, \ref{word:fold}) for the full FLORES-200 devtest.

Overall, all three encoders are generally able to retrieve semantically equivalent words in other languages from the same sentence. However, SALT more frequently retrieves tokens in non-Latin scripts and, more importantly, places true translations closer in terms of absolute cosine similarity. This is notable given our earlier geometric analysis, in which we observed that SALT yields a lower average cosine similarity across the embedding space than the other encoders.

Another improvement is that SALT reduces a common failure mode in which the model retrieves the same word in the same language from different sentential contexts. This suggests that SALT better preserves contextual meaning, rather than relying on surface-form similarity.

\begin{table}[t]
\centering
\small

\textbf{MEXMA}\\[4pt]
\begin{tabular}{r r r l l}
\toprule
Rank & Score & Lang & Unit \\
\midrule
1 & 0.945 & hun\_Latn & legtobb \\
2 & 0.943 & heb\_Hebr & rov \\
3 & 0.943 & nno\_Latn & fleste \\
4 & 0.942 & fra\_Latn & plupart \\
5 & 0.942 & ukr\_Cyrl & bilshosti \\
\bottomrule
\end{tabular}

\vspace{1em}

\textbf{SALT}\\[4pt]
\begin{tabular}{r r r l l}
\toprule
Rank & Score & Lang & Unit \\
\midrule
1 & 0.988 & ckb\_Arab & zorbey \\
2 & 0.987 & mar\_Deva & bahutek \\
3 & 0.986 & arz\_Arab & mu'zem \\
4 & 0.986 & npi\_Deva & adhikansh \\
5 & 0.984 & bul\_Cyrl & povecheto \\
\bottomrule
\end{tabular}

\vspace{1em}

\textbf{SONAR}\\[4pt]
\begin{tabular}{r r r l l}
\toprule
Rank & Score & Lang & Unit \\
\midrule
1 & 0.976 & fra\_Latn & plupart \\
2 & 0.971 & por\_Latn & maioria \\
3 & 0.971 & kor\_Hang & daebubun-ui \\
4 & 0.966 & spa\_Latn & mayoria \\
5 & 0.965 & cat\_Latn & majoria \\
\bottomrule
\end{tabular}

\caption{Nearest neighbours across models for the query word \textit{most} (all non-Latin scripts transliterated).}
\label{word:most}
\end{table}

\begin{table}[t]
\centering
\small

\textbf{MEXMA}\\[4pt]
\begin{tabular}{r r r l l}
\toprule
Rank & Score & Lang & Unit \\
\midrule
1 & 0.950 & ron\_Latn & insecte \\
2 & 0.940 & afr\_Latn & insekte \\
3 & 0.940 & spa\_Latn & insectos \\
4 & 0.939 & swe\_Latn & insekter \\
5 & 0.939 & epo\_Latn & insektoj \\
\bottomrule
\end{tabular}

\vspace{1em}

\textbf{SALT}\\[4pt]
\begin{tabular}{r r r l l}
\toprule
Rank & Score & Lang & Unit \\
\midrule
1 & 0.978 & afr\_Latn & insekte \\
2 & 0.970 & arz\_Arab & al-hasharat (al-ḥasharāt) \\
3 & 0.968 & slv\_Latn & insektov \\
4 & 0.967 & mkd\_Cyrl & insekti \\
5 & 0.967 & slk\_Latn & hmyzu \\
\bottomrule
\end{tabular}

\vspace{1em}

\textbf{SONAR}\\[4pt]
\begin{tabular}{r r r l l}
\toprule
Rank & Score & Lang & Unit \\
\midrule
1 & 0.970 & epo\_Latn & insektoj \\
2 & 0.964 & fra\_Latn & insectes \\
3 & 0.960 & nob\_Latn & insekter \\
4 & 0.954 & afr\_Latn & insekte \\
5 & 0.954 & deu\_Latn & Insekten \\
\bottomrule
\end{tabular}

\caption{Nearest neighbours for the query word \textit{insect} (all non-Latin scripts transliterated).}
\label{word:insect}
\end{table}

\begin{table}[t]
\centering
\small

\textbf{MEXMA}\\[4pt]
\begin{tabular}{r r l l r}
\toprule
Rank & Score & Lang & Unit & Sent. idx \\
\midrule
1 & 0.949 & eng\_Latn & fold & 621 \\
2 & 0.872 & tha\_Thai & a & 619 \\
3 & 0.873 & tha\_Thai & ph & 619 \\
4 & 0.873 & tha\_Thai & b & 619 \\
5 & 0.849 & tha\_Thai & a & 621 \\
\bottomrule
\end{tabular}

\vspace{1em}

\textbf{SALT}\\[4pt]
\begin{tabular}{r r l l r}
\toprule
Rank & Score & Lang & Unit & Sent. idx \\
\midrule
1 & 0.956 & ind\_Latn & melipat & 619 \\
2 & 0.946 & epo\_Latn & faldi & 619 \\
3 & 0.945 & ron\_Latn & plieze & 619 \\
4 & 0.937 & vie\_Latn & gap & 619 \\
5 & 0.936 & bul\_Cyrl & sgavat & 619 \\
\bottomrule
\end{tabular}

\vspace{1em}

\textbf{SONAR}\\[4pt]
\begin{tabular}{r r l l r}
\toprule
Rank & Score & Lang & Unit & Sent. idx \\
\midrule
1 & 0.915271 & epo\_Latn & faldi & 619 \\
2 & 0.901712 & kor\_Hang & jeobeul & 619 \\
3 & 0.874322 & heb\_Hebr & rov & 619 \\
4 & 0.874098 & eng\_Latn & fold & 621 \\
5 & 0.872106 & ind\_Latn & melipat & 619 \\
\bottomrule
\end{tabular}

\caption{Nearest neighbours across models for the query word \textit{fold} (all non-Latin scripts transliterated). Note that MEXMA and SONAR retrieve words from other sentences.}
\label{word:fold}
\end{table}

\section{Standard deviations}
\label{stdev}
Table~\ref{stdev:table} shows the standard deviation of our results from Table~\ref{main-results}.

\begin{table*}[]
\centering
\begin{tabular}{lccccc}
\toprule
\multicolumn{1}{r}{\textbf{}} & \textbf{AER} & \textbf{PAN-X} & \textbf{Massive} & \textbf{UPOS} & \textbf{WiC} \\
\midrule
SONAR Loss                    & 0.000255     & 0.000120       & 0.000622         & 0.000735       & 0.000361     \\
SO Loss                       & 0.000318     & 0.000311       & 0.001860         & 0.000445       & 0.000064     \\
WordOT                        & 0.002277     & 0.000516       & 0.001676         & 0.000156       & 0.000672     \\
WACSE                         & 0.000127     & 0.001146       & 0.000714         & 0.000113       & 0.001513     \\
OmniSONAR-Token               & 0.000068     & 0.000549       & 0.000630         & 0.000938       & 0.002037     \\
SALT                          & 0.000127     & 0.000456       & 0.001210         & 0.000677       & 0.000210    \\
\bottomrule
\end{tabular}
\caption{Standard deviation of our results in Table~\ref{main-results} (we run training experiments in three seeds).}
\label{stdev:table}
\end{table*}

\section{Human evaluation of spans}
\label{annotation}
For each of Chinese, French, Hindi, Russian, and Spanish, a native speaker annotated spans extracted from a total of 2,361 sentence pairs generated by either LLaMA-70B or CASE (with SONAR), without being informed of the originating system. The annotation split per language and system can be seen in Table~\ref{human_eval_results}. Annotators were presented with pairs of spans and asked to assess their semantic equivalence using a 4-point scale (see Appendix~\ref{rubric}). Further details on the recruitment process are provided in Appendix~\ref{recruitment}.

\subsection{Results on human evaluation of extracted aligned spans}
Table~\ref{human_eval_results} shows the score distribution of spans following human evaluation. We observe that, in almost all languages, spans of acceptable quality (Rating 3 or 4) occur at a rate higher than 90\%. These results validate using spans as a training signal. We also see that SONAR performance with CASE closely resembles LLM annotation.

\begin{table*}[]
\centering
\resizebox{0.8\textwidth}{!}{%
\begin{tabular}{cccccccc}

\toprule
\textbf{Language} & \textbf{Source} & \textbf{\#Annotations} & \textbf{Rating 1} & \textbf{Rating 2} & \textbf{Rating 3} & \textbf{Rating 4} & \textbf{Agg. 3+4} \\
\midrule
Spanish           & LLaMA-70B       & 129            & 0.0\%             & 3.1\%             & 0.8\%             & 96.1\%            & 96.9\%       \\
Spanish           & SONAR (CASE)    & 249            & 0.4\%             & 2.0\%             & 5.6\%             & 92.0\%            & 97.6\%       \\
Russian           & LLaMA-70B       & 102            & 0.0\%             & 7.8\%             & 20.6\%            & 71.6\%            & 92.2\%       \\
Russian           & SONAR (CASE)    & 204            & 3.4\%             & 8.3\%             & 21.1\%            & 67.2\%            & 88.3\%       \\
Hindi             & LLaMA-70B       & 334            & 3.9\%             & 7.2\%             & 15.3\%            & 73.7\%            & 89.0\%       \\
Hindi             & SONAR (CASE)    & 334            & 0.9\%             & 3.3\%             & 19.5\%            & 76.3\%            & 95.8\%       \\
French            & LLaMA-70B       & 135            & 0.0\%             & 5.2\%             & 11.1\%            & 83.7\%            & 94.8\%       \\
French            & SONAR (CASE)    & 264            & 0.4\%             & 8.0\%             & 21.6\%            & 70.1\%            & 91.7\%       \\
Chinese           & LLaMA-70B       & 280            & 1.8\%             & 5.4\%             & 13.2\%            & 79.6\%            & 92.8\%       \\
Chinese           & SONAR (CASE)    & 330            & 1.5\%             & 3.6\%             & 12.7\%            & 82.1\%            & 94.8\%      \\
\bottomrule
\end{tabular}}
\caption{Score distribution for the human evaluation of spans.}
\label{human_eval_results}
\end{table*}

\begin{table*}[]
\centering
\resizebox{1\textwidth}{!}{%
\begin{tabular}{cccccc}
\toprule
\textbf{Lang Pair} & \textbf{Source} & \textbf{en \#words (avg)} & \textbf{en \#words (median)} & \textbf{XX \#words (avg)} & \textbf{XX \#words (median)} \\
\midrule
English-Chinese    & LLaMA-70B       & 2.61                      & 2.0                          & 4.17                      & 3.5                          \\
English-Chinese    & SONAR (CASE)    & 5.70                      & 3.0                          & 8.42                      & 5.5                          \\
English-French     & LLaMA-70B       & 2.53                      & 2.0                          & 3.13                      & 3.0                          \\
English-French     & SONAR (CASE)    & 2.58                      & 2.0                          & 3.22                      & 2.0                          \\
English-Hindi      & LLaMA-70B       & 2.72                      & 2.0                          & 4.98                      & 4.0                          \\
English-Hindi      & SONAR (CASE)    & 2.82                      & 2.0                          & 5.70                      & 4.0                          \\
English-Russian    & LLaMA-70B       & 2.43                      & 2.0                          & 2.30                      & 2.0                          \\
English-Russian    & SONAR (CASE)    & 2.63                      & 2.0                          & 2.44                      & 2.0                          \\
English-Spanish    & LLaMA-70B       & 2.64                      & 2.0                          & 3.15                      & 3.0                          \\
English-Spanish    & SONAR (CASE)    & 2.43                      & 2.0                          & 2.59                      & 2.0                          \\
\midrule
All English-XX     & LLaMA-70B       & 2.62                      & 2.0                          & 3.97                      & 3.0                          \\
All English-XX     & SONAR (CASE)    & 3.36                      & 2.0                          & 4.83                      & 3.0                         \\ \bottomrule
\end{tabular}}
\caption{Average and median span word length.}
\label{table-span:stat}
\end{table*}

\begin{table*}[]
\resizebox{1\textwidth}{!}{%
\begin{tabular}{rrrrrrrr}
\toprule
\multicolumn{1}{c}{\textbf{Lang Pair}} & \multicolumn{1}{c}{\textbf{Source}} & \multicolumn{1}{c}{\textbf{1st POS}} & \multicolumn{1}{c}{\textbf{1st POS freq}} & \multicolumn{1}{c}{\textbf{2nd POS}} & \multicolumn{1}{c}{\textbf{2nd POS freq}} & \multicolumn{1}{c}{\textbf{3rd POS}} & \multicolumn{1}{c}{\textbf{3rd POS freq}} \\
\midrule
English-Chinese                        & LLaMA-70B                           & NOUN                                 & 55.4                                      & ADP                                  & 31.8                                      & VERB                                 & 26.8                                      \\
English-Chinese                        & SONAR (CASE)                        & NOUN                                 & 71.2                                      & ADP                                  & 51.2                                      & VERB                                 & 38.5                                      \\
English-French                         & LLaMA-70B                           & NOUN                                 & 54.1                                      & VERB                                 & 31.1                                      & ADP                                  & 26.7                                      \\
English-French                         & SONAR (CASE)                        & NOUN                                 & 49.6                                      & ADP                                  & 25.8                                      & VERB                                 & 23.9                                      \\
English-Hindi                          & LLaMA-70B                           & NOUN                                 & 48.8                                      & ADP                                  & 35.9                                      & VERB                                 & 29.6                                      \\
English-Hindi                          & SONAR (CASE)                        & NOUN                                 & 50.0                                      & DET                                  & 32.6                                      & PROPN                                & 28.4                                      \\
English-Russian                        & LLaMA-70B                           & NOUN                                 & 47.1                                      & VERB                                 & 38.2                                      & DET                                  & 25.5                                      \\
English-Russian                        & SONAR (CASE)                        & NOUN                                 & 56.9                                      & DET                                  & 27.9                                      & PRON                                 & 24.0                                      \\
English-Spanish                        & LLaMA-70B                           & NOUN                                 & 57.4                                      & ADP                                  & 29.5                                      & ADJ                                  & 27.1                                      \\
English-Spanish                        & SONAR (CASE)                        & NOUN                                 & 50.2                                      & VERB                                 & 26.9                                      & DET                                  & 26.1                                      \\
\midrule
All English-XX                         & LLaMA-70B                           & NOUN                                 & 52.3                                      & ADP                                  & 31.1                                      & VERB                                 & 29.5                                      \\
All English-XX                         & SONAR (CASE)                        & NOUN                                 & 56.0                                      & ADP                                  & 31.3                                      & DET                                  & 29.8             \\ \bottomrule                        
\end{tabular}}
\caption{Most common POS tags of English spans.}
\label{statistics_span:table}
\end{table*}

\subsection{Full instructions given to annotators}
\label{rubric}
Figure~\ref{instructions_participants} shows the annotation guidelines.

\subsection{Recruitment and demographics}
\label{recruitment}
Annotators were recruited through internal communication channels within our institution. All annotators were residing in the country where the institution is based at the time of the study. The annotator pool comprised seven individuals, each a native speaker of one of the following languages: Chinese, French, Hindi, Russian (two annotators) and Spanish (two annotators). 

Prior to the annotation process, all annotators participated in a training session in which we introduced the annotation interface, explained the evaluation rubric in detail, and jointly reviewed a set of example instances to ensure consistency in interpretation. Annotators were compensated at a rate of \$20.5 per hour.

\subsection{Participant Information Sheet}
Figure~\ref{fig:information_sheet} shows the Participant Information Sheet that was shared with the annotators. 

\section{Compute and model size}
\label{compute}
SONAR has 766M parameters, and SALT introduces no additional parameters when using the SONAR decoder. We do need a linear projection to adapt the XLM-Align representations to the SONAR decoder. We developed SALT using two NVIDIA A100 GPUs (80 GB each). Final training runs were conducted on two NVIDIA H200 GPUs (90 GB each) and required approximately two and a half days.
\paragraph{SALT relative to pre-training.}
SALT is a post-training procedure and is lightweight relative to pre-training.
\emph{Data:} we use $1.3$M sentence pairs (at most $40{,}000$ per language
pair), compared to the $43.7$B pairs of the NLLB dataset on which SONAR was
trained (over $30{,}000\times$ smaller). \emph{Compute:} SONAR was trained for
$100$k updates \citep{sonar}; assuming the effective batch of
roughly $1$M tokens per update used for NLLB \citep{nllb}, this
corresponds to approximately $100$B training tokens. SALT is trained for
$20$k updates with at most $1{,}400$ tokens per update, i.e., fewer than
$30$M tokens in total (over $3{,}500\times$ fewer), makes less than one pass
over its data, and introduces no new parameters. Both comparisons exclude span
extraction, a one-off offline step; the CASE heuristic removes the need for a
large LLM for it.

\section{Use of artifacts}
We use the SONAR encoder and the open-source subset of the NLLB Primary dataset, which are distributed under permissive licenses (Creative Commons Attribution Non-Commercial 4.0 and MIT License, respectively). These resources are used in accordance with their intended purpose of advancing research on multilingual sentence representations. All other encoders and datasets are used solely for evaluation, respecting their respective licenses and intended uses.

We do not apply additional filtering for offensive content or personally identifiable information (PII), as such filtering was already performed by the original dataset creators~\citep{nllb}.

\section{Usage of AI tools}
We acknowledge using AI tools for grammar correction and some other language clarifications, as well as code writing assistance.

\newpage

\begin{figure*}[h]
\begin{tcolorbox}[
  colback=gray!5,
  colframe=gray!40,
  title={\small\textbf{Prompt for LLM span alignment}},
  fonttitle=\small,
  left=4pt, right=4pt, top=4pt, bottom=4pt
]
\small\ttfamily
You are tasked with extracting aligned spans. You will be given two sentences, and need to identify spans of words/particles from the first one that appear in the second. Copy the spans of words as they appear originally in both texts. Only copy the extracted aligned spans, following the format of Example 1. If there are hallucinations or omissions, don't include them in the aligned spans.
\\
Example 1:

Sentence 1 [eng]: Samples of body fluids and tissues from people with the disease should be handled with special caution .

Sentence 2 [spa]: Las muestras de tejidos y fluidos corporales de personas con la enfermedad deben manejarse con especial precaución .
\\
Aligned spans:

[Samples] - [Las muestras]

[of body fluids] - [de fluidos corporales]

[and tissues] - [y tejidos]

[from people] - [de personas]

[with the disease] - [con la enfermedad]

[should be handled] - [deben manejarse]

[with special caution] - [con especial precaución]

Now, extract the alignments for the following pair:

Sentence 1 [\{src\_lang\}]: \{src\_text\}

Sentence 2 [\{tgt\_lang\}]: \{tgt\_text\}

Aligned spans:
\end{tcolorbox}
\caption{Prompt that was used on LLaMA-70B for span alignment.}
\label{prompt:span_llm}
\end{figure*}

\begin{table}[t]
\centering

\caption{Full list of languages used.}
\label{all_langs}
\footnotesize
\resizebox{1\columnwidth}{!}{%
\begin{tabular}{cccc}
\toprule
\textbf{Train} & & & (93 languages)\\
ace\_Arab & ace\_Latn & afr\_Latn & aka\_Latn \\
amh\_Ethi & arb\_Arab & ary\_Arab & arz\_Arab \\
bam\_Latn & ban\_Latn & ben\_Beng & bho\_Deva \\
bjn\_Arab & bjn\_Latn & bos\_Latn & bug\_Latn \\
bul\_Cyrl & crh\_Latn & dik\_Latn & dzo\_Tibt \\
eng\_Latn & ewe\_Latn & fon\_Latn & fra\_Latn \\
fur\_Latn & fuv\_Latn & gaz\_Latn & grn\_Latn \\
guj\_Gujr & hat\_Latn & hin\_Deva & hne\_Deva \\
hrv\_Latn & ind\_Latn & kan\_Knda & kas\_Arab \\
kas\_Deva & khm\_Khmr & kin\_Latn & knc\_Arab \\
knc\_Latn & lij\_Latn & lim\_Latn & lin\_Latn \\
lmo\_Latn & ltg\_Latn & lug\_Latn & mag\_Deva \\
mal\_Mlym & mar\_Deva & min\_Latn & mkd\_Cyrl \\
mni\_Beng & mri\_Latn & mya\_Mymr & nno\_Latn \\
nob\_Latn & nus\_Latn & ory\_Orya & pan\_Guru \\
pbt\_Arab & pes\_Arab & por\_Latn & prs\_Arab \\
rus\_Cyrl & scn\_Latn & shn\_Mymr & slv\_Latn \\
som\_Latn & sot\_Latn & spa\_Latn & srd\_Latn \\
srp\_Cyrl & ssw\_Latn & swh\_Latn & szl\_Latn \\
tam\_Taml & taq\_Latn & taq\_Tfng & tel\_Telu \\
tgl\_Latn & tir\_Ethi & tsn\_Latn & tso\_Latn \\
tzm\_Tfng & ukr\_Cyrl & urd\_Arab & vec\_Latn \\
xho\_Latn & yor\_Latn & zho\_Hans & zsm\_Latn \\
zul\_Latn \\
\midrule
\textbf{Test} & & & (57 languages) \\
afr\_Latn & als\_Latn & amh\_Ethi & arb\_Arab \\
azj\_Latn & ben\_Beng & bul\_Cyrl & cat\_Latn \\
cym\_Latn & dan\_Latn & deu\_Latn & ell\_Grek \\
eng\_Latn & est\_Latn & eus\_Latn & fin\_Latn \\
fra\_Latn & heb\_Hebr & hin\_Deva & hun\_Latn \\
hye\_Armn & ind\_Latn & isl\_Latn & ita\_Latn \\
jav\_Latn & jpn\_Jpan & kan\_Knda & kat\_Geor \\
kaz\_Cyrl & khk\_Cyrl & khm\_Khmr & kor\_Hang \\
lvs\_Latn & mal\_Mlym & mar\_Deva & mya\_Mymr \\
nld\_Latn & nob\_Latn & pes\_Arab & pol\_Latn \\
por\_Latn & ron\_Latn & rus\_Cyrl & slv\_Latn \\
spa\_Latn & swe\_Latn & swh\_Latn & tam\_Taml \\
tel\_Telu & tgl\_Latn & tha\_Thai & tur\_Latn \\
urd\_Arab & vie\_Latn & yor\_Latn & zho\_Hans \\
zsm\_Latn &           &           &            \\
\midrule
\textbf{Test $\setminus$ Train} & & & (29 languages) \\
als\_Latn & azj\_Latn & cat\_Latn & cym\_Latn \\
dan\_Latn & deu\_Latn & ell\_Grek & est\_Latn \\
eus\_Latn & fin\_Latn & heb\_Hebr & hun\_Latn \\
hye\_Armn & isl\_Latn & ita\_Latn & jav\_Latn \\
jpn\_Jpan & kat\_Geor & kaz\_Cyrl & khk\_Cyrl \\
kor\_Hang & lvs\_Latn & nld\_Latn & pol\_Latn \\
ron\_Latn & swe\_Latn & tha\_Thai & tur\_Latn \\
vie\_Latn & & & \\
\bottomrule

\end{tabular}}
\end{table}

\begin{figure*}[h]
\begin{tcolorbox}[
  colback=gray!5,
  colframe=gray!40,
  title={\small\textbf{Annotation instructions}},
  fonttitle=\small,
  left=4pt, right=4pt, top=4pt, bottom=4pt
]
\small\ttfamily

Purpose: evaluate the quality of automatically generated alignments between words or short phrases across different languages.

Why you were invited: you are fluent in Hindi, Russian, French, Chinese, or Spanish.

Participation: one remote session (about 120 minutes), voluntary, and withdrawal is possible up to two months after participation.

Data collected: Alignment judgments, response times, an anonymized annotator ID
\\

Explanation of Ratings
\\\\
\underline{\textbf{Rating 1 - Completely Misaligned}}

The spans are unrelated in meaning and do not correspond to each other. There is no meaningful semantic overlap.

\textbf{Example}

English: “These symptoms have been reported and [might] contribute to its rapid spreading”

Spanish: “Se han informado de esos síntomas, [lo que] podría contribuir a su rápida propagación”

Literal translation: “These symptoms have been reported, [what] may contribute to its rapid spreading”

\textbf{Explanation}: The spans (“might” vs. “lo que”) do not align in meaning or function.
\\\\
\underline{\textbf{Rating 2 - Partial Overlap, Core Meaning Mismatch}} 
The spans share some overlap, but fail to capture the main meaning or function of each other.

\textbf{Example 1}

English: “[Since cases] have so far clustered in certain regions in EU/EEA countries”

Spanish: “[Dado que, hasta el momento, se han formado conglomerados de casos] en ciertas regiones de países de la UE/EEE”

Literal translation: “[Since so far there have been clusters of cases] in certain regions of countries”

\textbf{Explanation}: There is overlap, but the English span (“Since cases”) is much shorter and does not capture the full meaning expressed in Spanish.

\textbf{Example 2}

English: “What [are the implications] for public health practice?”

Spanish: “¿Qué [consecuencias] tiene esto para la práctica de la salud pública?”

Literal translation: “What [consequences] this has for public health practice?”

\textbf{Explanation}: The Spanish span omits an important part of the meaning (the verbal structure “are / has”), so the alignment is incomplete.
\\\\
\underline{\textbf{Rating 3 - Same Role, Minor Differences}}

The spans fulfill the same semantic role and largely match in meaning, but there are small differences (e.g., missing words, slight shifts in nuance, or grammatical variation).

\textbf{Example 1}

English: “The virus is transmitted through respiratory droplets and [through objects].”

Spanish: “El virus se transmite por medio de las gotitas de la respiración y [los objetos].”

Literal translation: “The virus is transmitted through respiratory droplets and [the objects].”

\textbf{Explanation}: The meaning is almost identical, but the Spanish span lacks the preposition “through”.

\textbf{Example 2}

English: “Cabin crew should wear disposable medical gloves, [and possibly additional] personal protective equipment”

Spanish: “La tripulación de cabina debe usar guantes quirúrgicos desechables, [y, quizás,] equipos de protección individual”

Literal translation: “Cabin crew should wear disposable medical gloves, [and maybe] personal protective equipment”

\textbf{Explanation}: The meaning is very similar, but “additional” is not expressed in Spanish.
\\\\
\underline{\textbf{Rating 4 - Fully Aligned (Equivalent Meaning}}

The spans have the same meaning with no important differences. They could be swapped between sentences without changing the meaning.

\textbf{Example 1}

English: “[We are not aware] of any increased risk”

Spanish: “[No tenemos conocimiento] de un aumento de riesgo”

Literal translation: “[(We) do not have knowledge] of any increase in risk”

\textbf{Example 2}

English: “Particularly [avoid] cruise ship travel.”

Spanish: “En especial, [se deben evitar] los viajes en cruceros.”

Literal translation: “In particular, [they must be avoided] cruise ship travel”

\textbf{Explanation}: The spans are fully equivalent in meaning despite minor grammatical differences.
\end{tcolorbox}
\caption{Instructions shared with the annotators.}
\label{instructions_participants}
\end{figure*}

\begin{figure*}[h]
\begin{tcolorbox}[
  colback=gray!5,
  colframe=gray!40,
  title={\small\textbf{Participant Information Sheet (PIS)}},
  fonttitle=\small,
  left=4pt, right=4pt, top=4pt, bottom=4pt
]
\small\ttfamily

\textbf{What is the purpose of the study?}

The purpose of this study is to evaluate the quality of automatically generated alignments between words or short phrases across different languages (for example, “the dog” – “el perro”). These alignments are used in research on multilingual language technologies. By asking native speakers to judge whether these
alignments are correct, we aim to better understand how accurate these automatic methods are and improve techniques for learning cross-lingual language representations used in natural language processing systems.

\textbf{Why have I been asked to take part?}

You are a native speaker of the studied language (Russian, French, Chinese, Hindi or Spanish).

\textbf{Do I have to take part?}

No – participation in this study is entirely up to you. You can withdraw from the study at any time, up to 2 months, without giving a reason. After this point, it will no longer be possible to withdraw because we are not collecting any data that would allow us
to identify you.

\textbf{What will happen if I decide to take part?}
\begin{itemize}
    \item If you decide to take part, you will complete a short annotation task in which you review pairs of words or short phrases from two different languages (for example, “the dog” – “el perro”). You will be asked to indicate whether the alignment between the spans appears correct.
    \item The data collected will consist only of your judgments about the alignments and an anonymised annotator ID. The task will be completed through a simple online questionnaire or annotation interface.
    \item Participation will involve a single session lasting approximately 120 minutes. The task can be completed remotely at a time convenient for you.
\end{itemize}

\textbf{Compensation.}

Participants will receive financial compensation at a minimum rate of \$20.5 per hour for their time.

\textbf{Are there any risks associated with taking part?}

There are no significant risks associated with participation.

\textbf{What data are you collecting about me?}

The data we collect for our research is completely anonymous: We are not collecting any information that could, in our assessment, allow anyone to identify you. Your signed participant consent form will be kept separately from your responses and destroyed two months after your participation.

\textbf{What will happen to the results of this study?}

The results of this study may be summarised in published articles, reports and presentations. Your anonymised data may be published and can also be used for future research.

\textbf{Who can I contact?}

If you have any further questions about the study, please contact the lead researcher, [PLACEHOLDER]. If you wish to make a complaint about the study, please contact. [PLACEHOLDER]. When you contact us, please provide the study title and detail the nature of your complaint.

\textbf{Alternative formats.}

To request this document in an alternative format, such as large print or on coloured paper, please contact [PLACEHOLDER].

\end{tcolorbox}
\caption{Participant Information Sheet that was shared with the annotators.}
\label{fig:information_sheet}
\end{figure*}

\newpage

\newpage
\begin{figure*}
  \centering
\includegraphics[trim={0cm 0cm 0cm 32cm},clip, width=\linewidth]{latex/figs/seq_tagging_english_language_salt_vs_sonar.pdf}
    \caption{Performance of SALT vs SONAR on the sequence tagging tasks UPOS and WiC, on the English setting.}
    \label{per:lang:UPOS}
\end{figure*}

%% file: SALT/custom.bib
@inproceedings{fitzgerald2022massive,
    title = {{MASSIVE: A 1M-Example Multilingual Natural Language Understanding Dataset with 51 Typologically-Diverse Languages}},
    author = "FitzGerald, Jack  and
      Hench, Christopher  and
      Peris, Charith  and
      Mackie, Scott  and
      Rottmann, Kay  and
      Sanchez, Ana  and
      Nash, Aaron  and
      Urbach, Liam  and
      Kakarala, Vishesh  and
      Singh, Richa  and
      Ranganath, Swetha  and
      Crist, Laurie  and
      Britan, Misha  and
      Leeuwis, Wouter  and
      Tur, Gokhan  and
      Natarajan, Prem",
    editor = "Rogers, Anna  and
      Boyd-Graber, Jordan  and
      Okazaki, Naoaki",
    booktitle = "Proceedings of the 61st Annual Meeting of the Association for Computational Linguistics (Volume 1: Long Papers)",
    month = jul,
    year = "2023",
    address = "Toronto, Canada",
    publisher = "Association for Computational Linguistics",
    url = "https://aclanthology.org/2023.acl-long.235/",
    doi = "10.18653/v1/2023.acl-long.235",
    pages = "4277--4302"
}

@article{xtreme,
  author       = {Junjie Hu and
                  Sebastian Ruder and
                  Aditya Siddhant and
                  Graham Neubig and
                  Orhan Firat and
                  Melvin Johnson},
  title        = {{XTREME:} {A} Massively Multilingual Multi-task Benchmark for Evaluating
                  Cross-lingual Generalization},
  journal      = {CoRR},
  volume       = {abs/2003.11080},
  year         = {2020},
  url          = {https://arxiv.org/abs/2003.11080},
  eprinttype   = {arXiv},
  eprint       = {2003.11080},
  bibsource    = {dblp computer science bibliography, https://dblp.org}
}

@inproceedings{DBLP:conf/iclr/EnevoldsenCKKMS25,
  author       = {Kenneth C. Enevoldsen and
                  Isaac Chung and
                  Imene Kerboua and
                  M{\'{a}}rton Kardos and
                  Ashwin Mathur and
                  David Stap and
                  Jay Gala and
                  Wissam Siblini and
                  Dominik Krzeminski and
                  Genta Indra Winata and
                  Saba Sturua and
                  Saiteja Utpala and
                  Mathieu Ciancone and
                  Marion Schaeffer and
                  Diganta Misra and
                  Shreeya Dhakal and
                  Jonathan Rystr{\o}m and
                  Roman Solomatin and
                  {\"{O}}mer Veysel {\c{C}}agatan and
                  Akash Kundu and
                  et al.},
  title        = {{MMTEB:} Massive Multilingual Text Embedding Benchmark},
  booktitle    = {The Thirteenth International Conference on Learning Representations,
                  {ICLR} 2025, Singapore, April 24-28, 2025},
  publisher    = {OpenReview.net},
  year         = {2025},
  url          = {https://openreview.net/forum?id=zl3pfz4VCV},
  bibsource    = {dblp computer science bibliography, https://dblp.org}
}

@inproceedings{xlwa,
  author       = {Federico Martelli and
                  Andrei Stefan Bejgu and
                  Cesare Campagnano and
                  Jaka Cibej and
                  Rute Costa and
                  Apolonija Gantar and
                  Jelena Kallas and
                  Svetla Peneva Koeva and
                  Kristina Koppel and
                  Simon Krek and
                  Margit Langemets and
                  Veronika Lipp and
                  Sanni Nimb and
                  Sussi Olsen and
                  Bolette Sandford Pedersen and
                  Valeria Quochi and
                  Ana Salgado and
                  L{\'{a}}szl{\'{o}} Simon and
                  Carole Tiberius and
                  Rafael{-}J. Ure{\~{n}}a{-}Ruiz and
                  Roberto Navigli},
  editor       = {Federico Boschetti and
                  Gianluca E. Lebani and
                  Bernardo Magnini and
                  Nicole Novielli},
  title        = {{XL-WA:} a Gold Evaluation Benchmark for Word Alignment in 14 Language
                  Pairs},
  booktitle    = {Proceedings of the 9th Italian Conference on Computational Linguistics,
                  Venice, Italy, November 30 - December 2, 2023},
  series       = {{CEUR} Workshop Proceedings},
  publisher    = {CEUR-WS.org},
  year         = {2023},
  url          = {https://ceur-ws.org/Vol-3596/paper32.pdf},
  bibsource    = {dblp computer science bibliography, https://dblp.org}
}

@inproceedings{DBLP:conf/acl/KitaevCK19,
  author       = {Nikita Kitaev and
                  Steven Cao and
                  Dan Klein},
  editor       = {Anna Korhonen and
                  David R. Traum and
                  Llu{\'{\i}}s M{\`{a}}rquez},
  title        = {Multilingual Constituency Parsing with Self-Attention and Pre-Training},
  booktitle    = {Proceedings of the 57th Conference of the Association for Computational
                  Linguistics, {ACL} 2019, Florence, Italy, July 28- August 2, 2019,
                  Volume 1: Long Papers},
  pages        = {3499--3505},
  publisher    = {Association for Computational Linguistics},
  year         = {2019},
  url          = {https://doi.org/10.18653/v1/p19-1340},
  doi          = {10.18653/V1/P19-1340},
  bibsource    = {dblp computer science bibliography, https://dblp.org}
}

@inproceedings{conneau2018xnli,
  author = {Conneau, Alexis
and Rinott, Ruty
and Lample, Guillaume
and Williams, Adina
and Bowman, Samuel R.
and Schwenk, Holger
and Stoyanov, Veselin},
  booktitle = {{Proceedings of the 2018 Conference on Empirical Methods
in Natural Language Processing}},
  location = {Brussels, Belgium},
  publisher = {Association for Computational Linguistics},
  title = {{XNLI: Evaluating Cross-lingual Sentence Representations}},
  year = {2018},
}

@inproceedings{DBLP:conf/icml/0001I20,
  author       = {Tongzhou Wang and
                  Phillip Isola},
  title        = {Understanding Contrastive Representation Learning through Alignment
                  and Uniformity on the Hypersphere},
  booktitle    = {Proceedings of the 37th International Conference on Machine Learning,
                  {ICML} 2020, 13-18 July 2020, Virtual Event},
  series       = {Proceedings of Machine Learning Research},
  pages        = {9929--9939},
  publisher    = {{PMLR}},
  year         = {2020},
  url          = {http://proceedings.mlr.press/v119/wang20k.html},
  bibsource    = {dblp computer science bibliography, https://dblp.org}
}

@misc{neubig11kftt,
  author = {Graham Neubig},
  title = {The Kyoto Free Translation Task},
  howpublished = {http://www.phontron.com/kftt},
  year = {2011}
}

@inproceedings{DBLP:conf/acl/OchN00,
  author       = {Franz Josef Och and
                  Hermann Ney},
  title        = {Improved Statistical Alignment Models},
  booktitle    = {38th Annual Meeting of the Association for Computational Linguistics,
                  Hong Kong, China, October 1-8, 2000},
  pages        = {440--447},
  publisher    = {{ACL}},
  year         = {2000},
  url          = {https://aclanthology.org/P00-1056/},
  doi          = {10.3115/1075218.1075274},
  bibsource    = {dblp computer science bibliography, https://dblp.org}
}

@article{DBLP:journals/jmlr/RadovanovicNI10,
  author       = {Milos Radovanovic and
                  Alexandros Nanopoulos and
                  Mirjana Ivanovic},
  title        = {Hubs in Space: Popular Nearest Neighbors in High-Dimensional Data},
  journal      = {J. Mach. Learn. Res.},
  volume       = {11},
  pages        = {2487--2531},
  year         = {2010},
  url          = {https://dl.acm.org/doi/10.5555/1756006.1953015},
  doi          = {10.5555/1756006.1953015},
  bibsource    = {dblp computer science bibliography, https://dblp.org}
}

@inproceedings{xsimplusplus,
    title = {{xSIM++: An Improved Proxy to Bitext Mining Performance for Low-Resource Languages}},
    author = "Chen, Mingda  and
      Heffernan, Kevin  and
      {\c{C}}elebi, Onur  and
      Mourachko, Alexandre  and
      Schwenk, Holger",
    editor = "Rogers, Anna  and
      Boyd-Graber, Jordan  and
      Okazaki, Naoaki",
    booktitle = {{Proceedings of the 61st Annual Meeting of the Association for Computational Linguistics (Volume 2: Short Papers)}},
    month = jul,
    year = "2023",
    address = "Toronto, Canada",
    publisher = "Association for Computational Linguistics",
    url = "https://aclanthology.org/2023.acl-short.10/",
    doi = "10.18653/v1/2023.acl-short.10",
    pages = "101--109"
}

@inproceedings{muennighoff-etal-2023-mteb,
    title = "{MTEB}: Massive Text Embedding Benchmark",
    author = "Muennighoff, Niklas  and
      Tazi, Nouamane  and
      Magne, Loic  and
      Reimers, Nils",
    editor = "Vlachos, Andreas  and
      Augenstein, Isabelle",
    booktitle = "Proceedings of the 17th Conference of the European Chapter of the Association for Computational Linguistics",
    month = may,
    year = "2023",
    address = "Dubrovnik, Croatia",
    publisher = "Association for Computational Linguistics",
    url = "https://aclanthology.org/2023.eacl-main.148/",
    doi = "10.18653/v1/2023.eacl-main.148",
    pages = "2014--2037"
}

@inproceedings{DBLP:conf/eacl/DouN21,
  author       = {Zi{-}Yi Dou and
                  Graham Neubig},
  editor       = {Paola Merlo and
                  J{\"{o}}rg Tiedemann and
                  Reut Tsarfaty},
  title        = {Word Alignment by Fine-tuning Embeddings on Parallel Corpora},
  booktitle    = {Proceedings of the 16th Conference of the European Chapter of the
                  Association for Computational Linguistics: Main Volume, {EACL} 2021,
                  Online, April 19 - 23, 2021},
  pages        = {2112--2128},
  publisher    = {Association for Computational Linguistics},
  year         = {2021},
  url          = {https://doi.org/10.18653/v1/2021.eacl-main.181},
  doi          = {10.18653/V1/2021.EACL-MAIN.181},
  bibsource    = {dblp computer science bibliography, https://dblp.org}
}

@inproceedings{DBLP:conf/acl/AzadiFD23,
  author       = {Fatemeh Azadi and
                  Heshaam Faili and
                  Mohammad Javad Dousti},
  editor       = {Anna Rogers and
                  Jordan L. Boyd{-}Graber and
                  Naoaki Okazaki},
  title        = {PMI-Align: Word Alignment With Point-Wise Mutual Information Without
                  Requiring Parallel Training Data},
  booktitle    = {Findings of the Association for Computational Linguistics: {ACL} 2023,
                  Toronto, Canada, July 9-14, 2023},
  series       = {Findings of {ACL}},
  pages        = {12366--12377},
  publisher    = {Association for Computational Linguistics},
  year         = {2023},
  url          = {https://doi.org/10.18653/v1/2023.findings-acl.782},
  doi          = {10.18653/V1/2023.FINDINGS-ACL.782},
  bibsource    = {dblp computer science bibliography, https://dblp.org}
}

@inproceedings{labelproj,
  author       = {Tanmay Parekh and
                  I{-}Hung Hsu and
                  Kuan{-}Hao Huang and
                  Kai{-}Wei Chang and
                  Nanyun Peng},
  editor       = {Kevin Duh and
                  Helena G{\'{o}}mez{-}Adorno and
                  Steven Bethard},
  title        = {Contextual Label Projection for Cross-Lingual Structured Prediction},
  booktitle    = {Proceedings of the 2024 Conference of the North American Chapter of
                  the Association for Computational Linguistics: Human Language Technologies
                  (Volume 1: Long Papers), {NAACL} 2024, Mexico City, Mexico, June 16-21,
                  2024},
  pages        = {5738--5757},
  publisher    = {Association for Computational Linguistics},
  year         = {2024},
  url          = {https://doi.org/10.18653/v1/2024.naacl-long.321},
  doi          = {10.18653/V1/2024.NAACL-LONG.321},
  bibsource    = {dblp computer science bibliography, https://dblp.org}
}

@inproceedings{ottawa,
  author       = {Chenyang Huang and
                  Abbas Ghaddar and
                  Ivan Kobyzev and
                  Mehdi Rezagholizadeh and
                  Osmar Za{\"{\i}}ane and
                  Boxing Chen},
  editor       = {Lun{-}Wei Ku and
                  Andre Martins and
                  Vivek Srikumar},
  title        = {{OTTAWA:} Optimal TransporT Adaptive Word Aligner for Hallucination
                  and Omission Translation Errors Detection},
  booktitle    = {Findings of the Association for Computational Linguistics, {ACL} 2024,
                  Bangkok, Thailand and virtual meeting, August 11-16, 2024},
  series       = {Findings of {ACL}},
  pages        = {6322--6334},
  publisher    = {Association for Computational Linguistics},
  year         = {2024},
  url          = {https://doi.org/10.18653/v1/2024.findings-acl.377},
  doi          = {10.18653/V1/2024.FINDINGS-ACL.377},
  bibsource    = {dblp computer science bibliography, https://dblp.org}
}

@inproceedings{DBLP:conf/acl/TsiamasDC25,
  author       = {Ioannis Tsiamas and
                  David Dale and
                  Marta R. Costa{-}juss{\`{a}}},
  editor       = {Wanxiang Che and
                  Joyce Nabende and
                  Ekaterina Shutova and
                  Mohammad Taher Pilehvar},
  title        = {Improving Language and Modality Transfer in Translation by Character-level
                  Modeling},
  booktitle    = {Proceedings of the 63rd Annual Meeting of the Association for Computational
                  Linguistics (Volume 1: Long Papers), {ACL} 2025, Vienna, Austria,
                  July 27 - August 1, 2025},
  pages        = {20171--20187},
  publisher    = {Association for Computational Linguistics},
  year         = {2025},
  url          = {https://aclanthology.org/2025.acl-long.988/},
  bibsource    = {dblp computer science bibliography, https://dblp.org}
}

@article{nllb,
    title={{Scaling Neural Machine Translation to 200 Languages}}, 
    author={{NLLB Team}},
    year={2024},
    journal={Nature},
    volume={630}, 
    pages={841–846},
    doi={10.1038/s41586-024-07335-x},
    url={https://www.nature.com/articles/s41586-024-07335-x},
}

@inproceedings{DBLP:conf/emnlp/DaleC24,
  author       = {David Dale and
                  Marta R. Costa{-}juss{\`{a}}},
  editor       = {Yaser Al{-}Onaizan and
                  Mohit Bansal and
                  Yun{-}Nung Chen},
  title        = {{BLASER} 2.0: a metric for evaluation and quality estimation of massively
                  multilingual speech and text translation},
  booktitle    = {Findings of the Association for Computational Linguistics: {EMNLP}
                  2024, Miami, Florida, USA, November 12-16, 2024},
  series       = {Findings of {ACL}},
  pages        = {16075--16085},
  publisher    = {Association for Computational Linguistics},
  year         = {2024},
  url          = {https://doi.org/10.18653/v1/2024.findings-emnlp.943},
  doi          = {10.18653/V1/2024.FINDINGS-EMNLP.943},
  bibsource    = {dblp computer science bibliography, https://dblp.org}
}

@article{DBLP:journals/corr/abs-2207-04672,
  author       = {Marta R. Costa{-}juss{\`{a}} and
                  James Cross and
                  Onur {\c{C}}elebi and
                  Maha Elbayad and
                  Kenneth Heafield and
                  Kevin Heffernan and
                  Elahe Kalbassi and
                  Janice Lam and
                  Daniel Licht and
                  Jean Maillard and
                  Anna Y. Sun and
                  Skyler Wang and
                  Guillaume Wenzek and
                  Al Youngblood and
                  Bapi Akula and
                  Lo{\"{\i}}c Barrault and
                  Gabriel Mejia Gonzalez and
                  Prangthip Hansanti and
                  John Hoffman and
                  Semarley Jarrett and
                  Kaushik Ram Sadagopan and
                  Dirk Rowe and
                  Shannon Spruit and
                  Chau Tran and
                  Pierre Andrews and
                  Necip Fazil Ayan and
                  Shruti Bhosale and
                  Sergey Edunov and
                  Angela Fan and
                  Cynthia Gao and
                  Vedanuj Goswami and
                  Francisco Guzm{\'{a}}n and
                  Philipp Koehn and
                  Alexandre Mourachko and
                  Christophe Ropers and
                  Safiyyah Saleem and
                  Holger Schwenk and
                  Jeff Wang},
  title        = {No Language Left Behind: Scaling Human-Centered Machine Translation},
  journal      = {CoRR},
  volume       = {abs/2207.04672},
  year         = {2022},
  url          = {https://doi.org/10.48550/arXiv.2207.04672},
  doi          = {10.48550/ARXIV.2207.04672},
  eprinttype   = {arXiv},
  eprint       = {2207.04672},
  bibsource    = {dblp computer science bibliography, https://dblp.org}
}

@inproceedings{aer,
  author       = {Franz Josef Och and
                  Hermann Ney},
  title        = {Improved Statistical Alignment Models},
  booktitle    = {38th Annual Meeting of the Association for Computational Linguistics,
                  Hong Kong, China, October 1-8, 2000},
  pages        = {440--447},
  publisher    = {{ACL}},
  year         = {2000},
  url          = {https://aclanthology.org/P00-1056/},
  doi          = {10.3115/1075218.1075274},
  bibsource    = {dblp computer science bibliography, https://dblp.org}
}

@article{sonar2,
  author       = {{Omnilingual SONAR Team} and
                  Jo{\~a}o Maria Janeiro and
                  Pere-Llu{\'{\i}}s Huguet Cabot and
                  Ioannis Tsiamas and
                  Yen Meng and
                  Vivek Iyer and
                  Guillem Ram{\'{\i}}rez and
                  Loic Barrault and
                  Belen Alastruey and
                  Yu-An Chung and
                  Marta R. Costa-Jussa and
                  David Dale and
                  Kevin Heffernan and
                  Jaehyeong Jo and
                  Artyom Kozhevnikov and
                  Alexandre Mourachko and
                  Christophe Ropers and
                  Holger Schwenk and
                  Paul-Ambroise Duquenne},
  title        = {Omnilingual SONAR: Cross-Lingual and Cross-Modal Sentence Embeddings Bridging Massively Multilingual Text and Speech},
  journal      = {CoRR},
  volume       = {abs/2603.16606},
  year         = {2026},
  url          = {https://arxiv.org/abs/2603.16606},
  doi          = {10.48550/arXiv.2603.16606},
  eprinttype   = {arXiv},
  eprint       = {2603.16606},
  bibsource    = {dblp computer science bibliography, https://dblp.org}
}

@misc{omnilingualmtteam2026omnilingualmtmachinetranslation,
      title={Omnilingual MT: Machine Translation for 1,600 Languages}, 
      author={{Omnilingual MT Team} and Belen Alastruey and Niyati Bafna and Andrea Caciolai and Kevin Heffernan and Artyom Kozhevnikov and Christophe Ropers and Eduardo Sánchez and Charles-Eric Saint-James and Ioannis Tsiamas and Chierh Cheng and Joe Chuang and Paul-Ambroise Duquenne and Mark Duppenthaler and Nate Ekberg and Cynthia Gao and Pere Lluís Huguet Cabot and João Maria Janeiro and Jean Maillard and Gabriel Mejia Gonzalez and Holger Schwenk and Edan Toledo and Arina Turkatenko and Albert Ventayol-Boada and Rashel Moritz and Alexandre Mourachko and Surya Parimi and Mary Williamson and Shireen Yates and David Dale and Marta R. Costa-jussà},
      year={2026},
      eprint={2603.16309},
      archivePrefix={arXiv},
      primaryClass={cs.CL},
      url={https://arxiv.org/abs/2603.16309}, 
}

@inproceedings{DBLP:conf/lrec/NivreMGHMPSTZ20,
  author       = {Joakim Nivre and
                  Marie{-}Catherine de Marneffe and
                  Filip Ginter and
                  Jan Hajic and
                  Christopher D. Manning and
                  Sampo Pyysalo and
                  Sebastian Schuster and
                  Francis M. Tyers and
                  Daniel Zeman},
  editor       = {Nicoletta Calzolari and
                  Fr{\'{e}}d{\'{e}}ric B{\'{e}}chet and
                  Philippe Blache and
                  Khalid Choukri and
                  Christopher Cieri and
                  Thierry Declerck and
                  Sara Goggi and
                  Hitoshi Isahara and
                  Bente Maegaard and
                  Joseph Mariani and
                  H{\'{e}}l{\`{e}}ne Mazo and
                  Asunci{\'{o}}n Moreno and
                  Jan Odijk and
                  Stelios Piperidis},
  title        = {Universal Dependencies v2: An Evergrowing Multilingual Treebank Collection},
  booktitle    = {Proceedings of The 12th Language Resources and Evaluation Conference,
                  {LREC} 2020, Marseille, France, May 11-16, 2020},
  pages        = {4034--4043},
  publisher    = {European Language Resources Association},
  year         = {2020},
  url          = {https://aclanthology.org/2020.lrec-1.497/},
  bibsource    = {dblp computer science bibliography, https://dblp.org}
}

@inproceedings{DBLP:conf/naacl/PilehvarC19,
  author       = {Mohammad Taher Pilehvar and
                  Jos{\'{e}} Camacho{-}Collados},
  editor       = {Jill Burstein and
                  Christy Doran and
                  Thamar Solorio},
  title        = {WiC: the Word-in-Context Dataset for Evaluating Context-Sensitive
                  Meaning Representations},
  booktitle    = {Proceedings of the 2019 Conference of the North American Chapter of
                  the Association for Computational Linguistics: Human Language Technologies,
                  {NAACL-HLT} 2019, Minneapolis, MN, USA, June 2-7, 2019, Volume 1 (Long
                  and Short Papers)},
  pages        = {1267--1273},
  publisher    = {Association for Computational Linguistics},
  year         = {2019},
  url          = {https://doi.org/10.18653/v1/n19-1128},
  doi          = {10.18653/V1/N19-1128},
  bibsource    = {dblp computer science bibliography, https://dblp.org}
}

@inproceedings{DBLP:conf/acl/PanZMNKJ17,
  author       = {Xiaoman Pan and
                  Boliang Zhang and
                  Jonathan May and
                  Joel Nothman and
                  Kevin Knight and
                  Heng Ji},
  editor       = {Regina Barzilay and
                  Min{-}Yen Kan},
  title        = {Cross-lingual Name Tagging and Linking for 282 Languages},
  booktitle    = {Proceedings of the 55th Annual Meeting of the Association for Computational
                  Linguistics, {ACL} 2017, Vancouver, Canada, July 30 - August 4, Volume
                  1: Long Papers},
  pages        = {1946--1958},
  publisher    = {Association for Computational Linguistics},
  year         = {2017},
  url          = {https://doi.org/10.18653/v1/P17-1178},
  doi          = {10.18653/V1/P17-1178},
  bibsource    = {dblp computer science bibliography, https://dblp.org}
}

@inproceedings{alqahtani-etal-2021-using-optimal,
    title = "Using Optimal Transport as Alignment Objective for fine-tuning Multilingual Contextualized Embeddings",
    author = "Alqahtani, Sawsan  and
      Lalwani, Garima  and
      Zhang, Yi  and
      Romeo, Salvatore  and
      Mansour, Saab",
    editor = "Moens, Marie-Francine  and
      Huang, Xuanjing  and
      Specia, Lucia  and
      Yih, Scott Wen-tau",
    booktitle = "Findings of the Association for Computational Linguistics: EMNLP 2021",
    month = nov,
    year = "2021",
    address = "Punta Cana, Dominican Republic",
    publisher = "Association for Computational Linguistics",
    url = "https://aclanthology.org/2021.findings-emnlp.329/",
    doi = "10.18653/v1/2021.findings-emnlp.329",
    pages = "3904--3919"
}

@inproceedings{li-etal-2021-multi,
    title = "Multi-Granularity Contrasting for Cross-Lingual Pre-Training",
    author = "Li, Shicheng  and
      Yang, Pengcheng  and
      Luo, Fuli  and
      Xie, Jun",
    editor = "Zong, Chengqing  and
      Xia, Fei  and
      Li, Wenjie  and
      Navigli, Roberto",
    booktitle = "Findings of the Association for Computational Linguistics: ACL-IJCNLP 2021",
    month = aug,
    year = "2021",
    address = "Online",
    publisher = "Association for Computational Linguistics",
    url = "https://aclanthology.org/2021.findings-acl.149/",
    doi = "10.18653/v1/2021.findings-acl.149",
    pages = "1708--1717"
}

@inproceedings{dap,
    title = "Dual-Alignment Pre-training for Cross-lingual Sentence Embedding",
    author = "Li, Ziheng  and
      Huang, Shaohan  and
      Zhang, Zihan  and
      Deng, Zhi-Hong  and
      Lou, Qiang  and
      Huang, Haizhen  and
      Jiao, Jian  and
      Wei, Furu  and
      Deng, Weiwei  and
      Zhang, Qi",
    editor = "Rogers, Anna  and
      Boyd-Graber, Jordan  and
      Okazaki, Naoaki",
    booktitle = "Proceedings of the 61st Annual Meeting of the Association for Computational Linguistics (Volume 1: Long Papers)",
    month = jul,
    year = "2023",
    address = "Toronto, Canada",
    publisher = "Association for Computational Linguistics",
    url = "https://aclanthology.org/2023.acl-long.191/",
    doi = "10.18653/v1/2023.acl-long.191",
    pages = "3466--3478"
}

@article{DBLP:journals/corr/abs-2407-21783,
  author       = {Llama Team},
  title        = {The Llama 3 Herd of Models},
  journal      = {CoRR},
  volume       = {abs/2407.21783},
  year         = {2024},
  url          = {https://doi.org/10.48550/arXiv.2407.21783},
  doi          = {10.48550/ARXIV.2407.21783},
  eprinttype   = {arXiv},
  eprint       = {2407.21783},
  bibsource    = {dblp computer science bibliography, https://dblp.org}
}

@article{DBLP:journals/coling/BrownPPM94,
  author       = {Peter F. Brown and
                  Stephen Della Pietra and
                  Vincent J. Della Pietra and
                  Robert L. Mercer},
  title        = {The Mathematics of Statistical Machine Translation: Parameter Estimation},
  journal      = {Comput. Linguistics},
  volume       = {19},
  number       = {2},
  pages        = {263--311},
  year         = {1993},
  bibsource    = {dblp computer science bibliography, https://dblp.org}
}

@inproceedings{DBLP:conf/naacl/KoehnOM03,
  author       = {Philipp Koehn and
                  Franz Josef Och and
                  Daniel Marcu},
  editor       = {Marti A. Hearst and
                  Mari Ostendorf},
  title        = {Statistical Phrase-Based Translation},
  booktitle    = {Human Language Technology Conference of the North American Chapter
                  of the Association for Computational Linguistics, {HLT-NAACL} 2003,
                  Edmonton, Canada, May 27 - June 1, 2003},
  publisher    = {The Association for Computational Linguistics},
  year         = {2003},
  url          = {https://aclanthology.org/N03-1017/},
  bibsource    = {dblp computer science bibliography, https://dblp.org}
}

@inproceedings{chi-etal-2021-improving,
    title = "Improving Pretrained Cross-Lingual Language Models via Self-Labeled Word Alignment",
    author = "Chi, Zewen  and
      Dong, Li  and
      Zheng, Bo  and
      Huang, Shaohan  and
      Mao, Xian-Ling  and
      Huang, Heyan  and
      Wei, Furu",
    editor = "Zong, Chengqing  and
      Xia, Fei  and
      Li, Wenjie  and
      Navigli, Roberto",
    booktitle = "Proceedings of the 59th Annual Meeting of the Association for Computational Linguistics and the 11th International Joint Conference on Natural Language Processing (Volume 1: Long Papers)",
    month = aug,
    year = "2021",
    address = "Online",
    publisher = "Association for Computational Linguistics",
    url = "https://aclanthology.org/2021.acl-long.265/",
    doi = "10.18653/v1/2021.acl-long.265",
    pages = "3418--3430"
}

@inproceedings{DBLP:conf/iclr/WeiW0XYL21,
  author       = {Xiangpeng Wei and
                  Rongxiang Weng and
                  Yue Hu and
                  Luxi Xing and
                  Heng Yu and
                  Weihua Luo},
  title        = {On Learning Universal Representations Across Languages},
  booktitle    = {9th International Conference on Learning Representations, {ICLR} 2021,
                  Virtual Event, Austria, May 3-7, 2021},
  publisher    = {OpenReview.net},
  year         = {2021},
  url          = {https://openreview.net/forum?id=Uu1Nw-eeTxJ},
  bibsource    = {dblp computer science bibliography, https://dblp.org}
}

@inproceedings{miao-etal-2024-enhancing,
    title = "Enhancing Cross-lingual Sentence Embedding for Low-resource Languages with Word Alignment",
    author = "Miao, Zhongtao  and
      Wu, Qiyu  and
      Zhao, Kaiyan  and
      Wu, Zilong  and
      Tsuruoka, Yoshimasa",
    editor = "Duh, Kevin  and
      Gomez, Helena  and
      Bethard, Steven",
    booktitle = "Findings of the Association for Computational Linguistics: NAACL 2024",
    month = jun,
    year = "2024",
    address = "Mexico City, Mexico",
    publisher = "Association for Computational Linguistics",
    url = "https://aclanthology.org/2024.findings-naacl.204/",
    doi = "10.18653/v1/2024.findings-naacl.204",
    pages = "3225--3236"
}

@inproceedings{mehta-varma-2023-llm,
    title = "{LLM}-{RM} at {S}em{E}val-2023 Task 2: Multilingual Complex {NER} Using {XLM}-{R}o{BERT}a",
    author = "Mehta, Rahul  and
      Varma, Vasudeva",
    editor = {Ojha, Atul Kr.  and
      Do{\u{g}}ru{\"o}z, A. Seza  and
      Da San Martino, Giovanni  and
      Tayyar Madabushi, Harish  and
      Kumar, Ritesh  and
      Sartori, Elisa},
    booktitle = "Proceedings of the 17th International Workshop on Semantic Evaluation (SemEval-2023)",
    month = jul,
    year = "2023",
    address = "Toronto, Canada",
    publisher = "Association for Computational Linguistics",
    url = "https://aclanthology.org/2023.semeval-1.62/",
    doi = "10.18653/v1/2023.semeval-1.62",
    pages = "453--456"
}

@inproceedings{simalign,
    title = "{S}im{A}lign: High Quality Word Alignments Without Parallel Training Data Using Static and Contextualized Embeddings",
    author = {Jalili Sabet, Masoud  and
      Dufter, Philipp  and
      Yvon, Fran{\c{c}}ois  and
      Sch{\"u}tze, Hinrich},
    editor = "Cohn, Trevor  and
      He, Yulan  and
      Liu, Yang",
    booktitle = "Findings of the Association for Computational Linguistics: EMNLP 2020",
    month = nov,
    year = "2020",
    address = "Online",
    publisher = "Association for Computational Linguistics",
    url = "https://aclanthology.org/2020.findings-emnlp.147/",
    doi = "10.18653/v1/2020.findings-emnlp.147",
    pages = "1627--1643"
}

@inproceedings{xlm,
  author       = {Alexis Conneau and
                  Guillaume Lample},
  editor       = {Hanna M. Wallach and
                  Hugo Larochelle and
                  Alina Beygelzimer and
                  Florence d'Alch{\'{e}}{-}Buc and
                  Emily B. Fox and
                  Roman Garnett},
  title        = {Cross-lingual Language Model Pretraining},
  booktitle    = {Advances in Neural Information Processing Systems 32: Annual Conference
                  on Neural Information Processing Systems 2019, NeurIPS 2019, December
                  8-14, 2019, Vancouver, BC, Canada},
  pages        = {7057--7067},
  year         = {2019},
  url          = {https://proceedings.neurips.cc/paper/2019/hash/c04c19c2c2474dbf5f7ac4372c5b9af1-Abstract.html},
  bibsource    = {dblp computer science bibliography, https://dblp.org}
}

@inproceedings{DBLP:conf/iclr/LoshchilovH19,
  author       = {Ilya Loshchilov and
                  Frank Hutter},
  title        = {Decoupled Weight Decay Regularization},
  booktitle    = {7th International Conference on Learning Representations, {ICLR} 2019,
                  New Orleans, LA, USA, May 6-9, 2019},
  publisher    = {OpenReview.net},
  year         = {2019},
  url          = {https://openreview.net/forum?id=Bkg6RiCqY7},
  bibsource    = {dblp computer science bibliography, https://dblp.org}
}

@inproceedings{labse,
  author       = {Fangxiaoyu Feng and
                  Yinfei Yang and
                  Daniel Cer and
                  Naveen Arivazhagan and
                  Wei Wang},
  editor       = {Smaranda Muresan and
                  Preslav Nakov and
                  Aline Villavicencio},
  title        = {Language-agnostic {BERT} Sentence Embedding},
  booktitle    = {Proceedings of the 60th Annual Meeting of the Association for Computational
                  Linguistics (Volume 1: Long Papers), {ACL} 2022, Dublin, Ireland,
                  May 22-27, 2022},
  pages        = {878--891},
  publisher    = {Association for Computational Linguistics},
  year         = {2022},
  url          = {https://doi.org/10.18653/v1/2022.acl-long.62},
  doi          = {10.18653/V1/2022.ACL-LONG.62},
  bibsource    = {dblp computer science bibliography, https://dblp.org}
}

@article{sonar,
  author       = {Paul{-}Ambroise Duquenne and
                  Holger Schwenk and
                  Beno{\^{\i}}t Sagot},
  title        = {{SONAR:} Sentence-Level Multimodal and Language-Agnostic Representations},
  journal      = {CoRR},
  volume       = {abs/2308.11466},
  year         = {2023},
  url          = {https://doi.org/10.48550/arXiv.2308.11466},
  doi          = {10.48550/ARXIV.2308.11466},
  eprinttype    = {arXiv},
  eprint       = {2308.11466},
  bibsource    = {dblp computer science bibliography, https://dblp.org}
}

@inproceedings{mexma,
  author       = {Jo{\~{a}}o Maria Janeiro and
                  Benjamin Piwowarski and
                  Patrick Gallinari and
                  Lo{\"{\i}}c Barrault},
  editor       = {Wanxiang Che and
                  Joyce Nabende and
                  Ekaterina Shutova and
                  Mohammad Taher Pilehvar},
  title        = {{MEXMA:} Token-level objectives improve sentence representations},
  booktitle    = {Proceedings of the 63rd Annual Meeting of the Association for Computational
                  Linguistics (Volume 1: Long Papers), {ACL} 2025, Vienna, Austria,
                  July 27 - August 1, 2025},
  pages        = {23960--23995},
  publisher    = {Association for Computational Linguistics},
  year         = {2025},
  url          = {https://aclanthology.org/2025.acl-long.1168/},
  bibsource    = {dblp computer science bibliography, https://dblp.org}
}


%% file: custom.bib
@article{adelani-etal-2021-masakhaner,
    title = "{M}asakha{NER}: Named Entity Recognition for {A}frican Languages",
    author = "Adelani, David Ifeoluwa  and
      Abbott, Jade  and
      Neubig, Graham  and
      D{'}souza, Daniel  and
      Kreutzer, Julia  and
      Lignos, Constantine  and
      Palen-Michel, Chester  and
      Buzaaba, Happy  and
      Rijhwani, Shruti  and
      Ruder, Sebastian  and
      Mayhew, Stephen  and
      Azime, Israel Abebe  and
      Muhammad, Shamsuddeen H.  and
      Emezue, Chris Chinenye  and
      Nakatumba-Nabende, Joyce  and
      Ogayo, Perez  and
      Anuoluwapo, Aremu  and
      Gitau, Catherine  and
      Mbaye, Derguene  and
      Alabi, Jesujoba  and
      Yimam, Seid Muhie  and
      Gwadabe, Tajuddeen Rabiu  and
      Ezeani, Ignatius  and
      Niyongabo, Rubungo Andre  and
      Mukiibi, Jonathan  and
      Otiende, Verrah  and
      Orife, Iroro  and
      David, Davis  and
      Ngom, Samba  and
      Adewumi, Tosin  and
      Rayson, Paul  and
      Adeyemi, Mofetoluwa  and
      Muriuki, Gerald  and
      Anebi, Emmanuel  and
      Chukwuneke, Chiamaka  and
      Odu, Nkiruka  and
      Wairagala, Eric Peter  and
      Oyerinde, Samuel  and
      Siro, Clemencia  and
      Bateesa, Tobius Saul  and
      Oloyede, Temilola  and
      Wambui, Yvonne  and
      Akinode, Victor  and
      Nabagereka, Deborah  and
      Katusiime, Maurice  and
      Awokoya, Ayodele  and
      MBOUP, Mouhamadane  and
      Gebreyohannes, Dibora  and
      Tilaye, Henok  and
      Nwaike, Kelechi  and
      Wolde, Degaga  and
      Faye, Abdoulaye  and
      Sibanda, Blessing  and
      Ahia, Orevaoghene  and
      Dossou, Bonaventure F. P.  and
      Ogueji, Kelechi  and
      DIOP, Thierno Ibrahima  and
      Diallo, Abdoulaye  and
      Akinfaderin, Adewale  and
      Marengereke, Tendai  and
      Osei, Salomey",
    editor = "Roark, Brian  and
      Nenkova, Ani",
    journal = "Transactions of the Association for Computational Linguistics",
    volume = "9",
    year = "2021",
    address = "Cambridge, MA",
    publisher = "MIT Press",
    url = "https://aclanthology.org/2021.tacl-1.66/",
    doi = "10.1162/tacl_a_00416",
    pages = "1116--1131"
}

@inproceedings{garcia-ferrero-etal-2022-model,
    title = "Model and Data Transfer for Cross-Lingual Sequence Labelling in Zero-Resource Settings",
    author = "Garc{\'i}a-Ferrero, Iker  and
      Agerri, Rodrigo  and
      Rigau, German",
    editor = "Goldberg, Yoav  and
      Kozareva, Zornitsa  and
      Zhang, Yue",
    booktitle = "Findings of the Association for Computational Linguistics: EMNLP 2022",
    month = dec,
    year = "2022",
    address = "Abu Dhabi, United Arab Emirates",
    publisher = "Association for Computational Linguistics",
    url = "https://aclanthology.org/2022.findings-emnlp.478/",
    doi = "10.18653/v1/2022.findings-emnlp.478",
    pages = "6403--6416"
}

@inproceedings{yarmohammadi-etal-2021-everything,
    title = "Everything Is All It Takes: A Multipronged Strategy for Zero-Shot Cross-Lingual Information Extraction",
    author = "Yarmohammadi, Mahsa  and
      Wu, Shijie  and
      Marone, Marc  and
      Xu, Haoran  and
      Ebner, Seth  and
      Qin, Guanghui  and
      Chen, Yunmo  and
      Guo, Jialiang  and
      Harman, Craig  and
      Murray, Kenton  and
      White, Aaron Steven  and
      Dredze, Mark  and
      Van Durme, Benjamin",
    editor = "Moens, Marie-Francine  and
      Huang, Xuanjing  and
      Specia, Lucia  and
      Yih, Scott Wen-tau",
    booktitle = "Proceedings of the 2021 Conference on Empirical Methods in Natural Language Processing",
    month = nov,
    year = "2021",
    address = "Online and Punta Cana, Dominican Republic",
    publisher = "Association for Computational Linguistics",
    url = "https://aclanthology.org/2021.emnlp-main.149/",
    doi = "10.18653/v1/2021.emnlp-main.149",
    pages = "1950--1967"
}

@inproceedings{dale-etal-2023-halomi,
    title = "{H}al{O}mi: A Manually Annotated Benchmark for Multilingual Hallucination and Omission Detection in Machine Translation",
    author = {Dale, David  and
      Voita, Elena  and
      Lam, Janice  and
      Hansanti, Prangthip  and
      Ropers, Christophe  and
      Kalbassi, Elahe  and
      Gao, Cynthia  and
      Barrault, Lo{\"i}c  and
      Costa-juss{\`a}, Marta R.},
    editor = "Bouamor, Houda  and
      Pino, Juan  and
      Bali, Kalika",
    booktitle = "Proceedings of the 2023 Conference on Empirical Methods in Natural Language Processing",
    month = dec,
    year = "2023",
    address = "Singapore",
    publisher = "Association for Computational Linguistics",
    url = "https://aclanthology.org/2023.emnlp-main.42/",
    doi = "10.18653/v1/2023.emnlp-main.42",
    pages = "638--653"
}
